\documentclass[10pt,twocolumn,letterpaper]{article}

\usepackage{cvpr}              % To produce the CAMERA-READY version

\newcommand{\s}{\boldsymbol{s}}

\newcommand{\z}{\boldsymbol{z}}
\newcommand{\p}{\boldsymbol{p}}

\newcommand{\vel}{\boldsymbol{v}}
\newcommand{\V}{\mathbb{V}}
\newcommand{\Dt}{\Delta t}
\newcommand{\R}{\mathcal{R}}
\usepackage{multirow}
\usepackage{ulem}
\usepackage{algpseudocode}
\usepackage{algorithm}
\usepackage{wrapfig}
\usepackage{comment}
\usepackage{pifont}

\newcommand{\N}{\ding{55}}

\newcommand{\myrightleftarrows}[1]{\mathrel{\substack{\xrightarrow{#1} \\[-.9ex] \xleftarrow{#1}}}}

\definecolor{cvprblue}{rgb}{0.21,0.49,0.74}
\usepackage[pagebackref,breaklinks,colorlinks,allcolors=cvprblue]{hyperref}

\def\paperID{1138} % *** Enter the Paper ID here
\def\confName{CVPR}
\def\confYear{2025}

\newcommand{\nickname}{FreeGave}
\begin{document}

\title{\nickname{}: 3D Physics Learning from Dynamic Videos by Gaussian Velocity}

%%%%%%%%% AUTHORS - PLEASE UPDATE
\author{Jinxi Li, \quad Ziyang Song, \quad Siyuan Zhou, \quad Bo Yang\thanks{Corresponding Author}\\
vLAR Group, The Hong Kong Polytechnic University\\
{\tt\small \{jinxi.li, ziyang.song, siyuan.zhou\}@connect.polyu.hk, bo.yang@polyu.edu.hk}}
% \and
% \maketitle

\twocolumn[{%
\renewcommand\twocolumn[1][]{#1}%
    \maketitle
    \begin{center}
        \vspace{-15pt}
        \centering
        \includegraphics[scale=0.43]{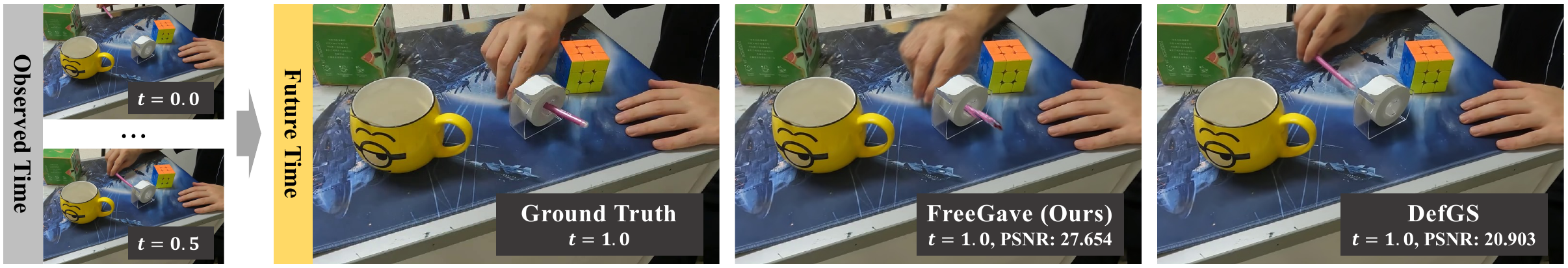}
        \vspace{-15pt}
        \captionof{figure}{Given video frames of a real-world dynamic scene, our \nickname{} can learn the underlying physics and  accurately predict the future motion of a purple pen going through the hole of a tape, while existing baselines cannot.}
        \label{fig:opening}
        \vspace{0pt}
    \end{center}
}]

% \renewcommand{\thefootnote}{\fnsymbol{footnote}}
% \footnote[1]{Corresponding author}

\begin{abstract}
\indent In this paper, we aim to model 3D scene geometry, appearance, and the underlying physics purely from multi-view videos. By applying various governing PDEs as PINN losses or incorporating physics simulation into neural networks, existing works often fail to learn complex physical motions at boundaries or require object priors such as masks or types. In this paper, we propose \textbf{\nickname{}} to learn physics of complex dynamic 3D scenes without needing any object priors. The key to our approach is to introduce a physics code followed by a carefully designed divergence-free module for estimating a per-Gaussian velocity field, without relying on the inefficient PINN losses. Extensive experiments on three public datasets and a newly collected challenging real-world dataset demonstrate the superior performance of our method for future frame extrapolation and motion segmentation. Most notably, our investigation into the learned physics codes reveals that they truly learn meaningful 3D physical motion patterns in the absence of any human labels in training. Our code and data are available at {\small\url{https://github.com/vLAR-group/FreeGave}}

\end{abstract}   \vspace{-0.5cm}
\section{Introduction} \label{sec:intro}
\phantom{xx} Obtaining an accurate physical model for complex dynamic 3D environments is a key enabler for many emerging applications, including robot planning, embodied interaction, and dynamic 3D/4D generation. Represented by NeRF \cite{Mildenhall2020}, 3DGS \cite{Kerbl2023} and their variants \cite{Pumarola2021,Barron2021a,Li2023b,Yang2024,Wu2024,Kratimenos2024}, recent advancements in 3D learning have demonstrated an unprecedented level of fidelity in modeling dynamic 3D scenes, achieving excellent performance in novel view interpolation and/or 3D shape reconstruction. However, the majority of these approaches lack the core ability of future prediction, essentially because they fail to effectively learn physical properties of 3D space.

To learn physical properties, one line of works introduces various governing PDEs, commonly known as PINN \cite{Raissi2019}, into the learning process of dynamic 3D scenes. While achieving encouraging results in modeling 3D physics such as velocity and viscosity \cite{Chu2022,Li2023c,Li2023,Wang2024}, the integration of PINN losses often compromises the accuracy of learned physics in boundary regions, leading to blurred predictions for future frames. Another line of works explicitly incorporates various physics models into neural networks \cite{Jonathan2020,Whitney2024,Zhong2024,Cai2024}, showing impressive results in learning physical properties and further transferring to simulation. Nevertheless, these methods are usually limited to specific types of objects or motions, as the encoded physics models are often less generic than PINN constraints, necessitating additional boundary conditions such as object masks. 

In this work, we aim to present a simple and general method for learning physics of complex dynamic 3D scenes using only multi-view videos, without needing any PINN losses or extra priors such as object types or masks, which are often needed by existing methods. Among various physical properties of a dynamic 3D scene, we follow the recent work NVFi \cite{Li2023c} and focus on learning 3D velocity as it directly governs 3D dynamics and future predictions. 

Given video frames containing an unknown number of objects/parts exhibiting diverse motions, at a single glance, it is extremely challenging to learn velocities of all (unknown) objects or parts in 3D space, due to the lack of physics and semantics in raw pixels. Nevertheless, the whole dynamic 3D scene can be simply regarded as a collection of many 3D rigid particles with their own sizes, orientations, and motion patterns. The dynamics of each particle are actually governed by its hidden physical parameters, such as mass and forces. Thus, our challenge of learning complex 3D scene dynamics can be naturally framed as learning dynamics of these 3D particles in general. 

With this motivation, we propose an elegant pipeline to simultaneously learn 3D geometry, appearance, and velocity of dynamic scenes. Treating each 3D Gaussian kernel as a rigid particle, we extend the successful 3DGS \cite{Kerbl2023} to learn an additional physics code for each 3D Gaussian. With such a code, the 3D velocity of each Gaussian is then learned from visual pixels without relying on PINN losses. Particularly, our pipeline comprises three major components: 1) a \textbf{canonical 3D representation module} which adopts a vanilla 3DGS to learn the dynamic 3D scene geometry and appearance at a canonical timestamp; 2) a \textbf{neural divergence-free Gaussian velocity module} to estimate per-Gaussian velocity based on the introduced latent physics code; and 3) a \textbf{deformation-aided optimization module} to drive the velocity to be physically meaningful. 

The core to our pipeline lies in the latter two modules. Basically, given video frames of a dynamic 3D scene, our neural divergence-\textbf{free} \textbf{Ga}ussian \textbf{ve}locity module together with the optimization strategy explicitly drives the per-Gaussian physics code to describe the corresponding motion patterns in latent space. This ensures that the estimated velocities adhere to fundamental physics rules, ultimately enabling precise predictions of future frames. Figure \ref{fig:opening} shows qualitative results of a challenging dynamic scene. Our method is named \textbf{\nickname{}} and our contributions are:
\begin{itemize}[leftmargin=*]
\setlength{\itemsep}{1pt}
\setlength{\parsep}{2pt}
\setlength{\parskip}{1pt}
    \item We introduce a general framework for learning physics of dynamic 3D scenes solely from RGB videos, eliminating the need for any object priors, such as types and masks.  
    \item We propose to learn per-Gaussian velocities from latent physics codes, guided by a carefully designed divergence-free ingredient instead of inefficient PINN losses.   
    \item We showcase superior results in future frame extrapolation on three existing datasets, and a newly collected real-world dataset featuring highly challenging dynamics. 
\end{itemize}

\section{Related Works}\label{sec:liter}

\phantom{xx}\textbf{3D Representation Learning}: Classic methods for representing 3D geometry mainly include point clouds \cite{Fan2017}, meshes \cite{Kato2017}, voxels \cite{Yang2018}, and primitives \cite{Zou2017}. However, the fidelity of 3D shapes is often limited by the spatial resolution inherent in these explicit representations. To address this limitation, a line of neural 3D representations has been introduced recently, including un/signed distance fields \cite{Chibane2020a,Wang2022,Park2019}, occupancy fields \cite{Chen2019g,Mescheder2019}, and radiance fields \cite{Mildenhall2020}. Leveraging coordinate-based MLPs, these representations have demonstrated remarkable performance in novel view rendering and 3D shape recovery \cite{Trevithick2021,Xie2022,Tewari2021}. Nonetheless, the integration of coordinate-based representations leads to slow extraction of 2D views and 3D surfaces. To resolve this issue, 3DGS \cite{Kerbl2023} is introduced to represent a 3D shape as a set of 3D Gaussian kernels with various properties, achieving real-time rendering speeds as well as exceptional quality in novel view synthesis \cite{Fei2024,Chen2024,Bao2024}. In our pipeline, we adopt 3DGS as our 3D shape and appearance representation, treating each Gaussian kernel as a rigid particle for learning its physics code and velocity. 

\textbf{Dynamic 3D Reconstruction}: To reconstruct dynamic 3D objects and scenes, existing works are primarily built on established static 3D representations such as SDF \cite{Park2019}, NeRF \cite{Mildenhall2020}, and 3DGS \cite{Kerbl2023}. Given dynamic visual input over time, current methods \cite{Park2021,Tretschk2021,Gao2021,Tian2023,Liu2023,Fridovich-Keil2023,Cao2023a,Liu2022,Fang2022,Yao2024,Yang2024a,Im2024,Zhan2024,Song2024} commonly add the time dimension into a static 3D representation to learn the underlying motion for rigid or deformable targets. While obtaining impressive accuracy in novel view rendering, especially when utilizing 3DGS as the 3D representation \cite{Lin2024,Lu2024,Li2024,Qian2024,Wu2024,Shaw2024,Mihajlovic2024,Bae2024,Yang2024b,Katsumata2024}, these methods can only synthesize high-quality views within the observed time, lacking the ability to predict physically meaningful future frames. This limitation arises because these methods fail to encode physics priors into the learning process, but just fit the visual observations during training. In our pipeline, we explicitly learn a physics code for each Gaussian via our Gaussian velocity module and the divergence-free parameterization. 

\textbf{Video Prediction}: Some studies focus on directly predicting future video frames \cite{Fragkiadaki2016,Hafner2019}. Thanks to the powerful diffusion model and Transformer-based architectures, many of these methods \cite{Hoppe2022,Ho2022,Voleti2022,Ni2023,Yu2023,Wu2023,Chen2024a,Liu2024} have demonstrated superior performance in video generation. However, they are prone to predicting physically implausible motions, primarily because their networks do not explicitly incorporate physics priors in training, as discussed in \cite{Liu2024b,Kang2024}.  
\begin{figure*}[t]
\centering
\includegraphics[width=1.0\linewidth]{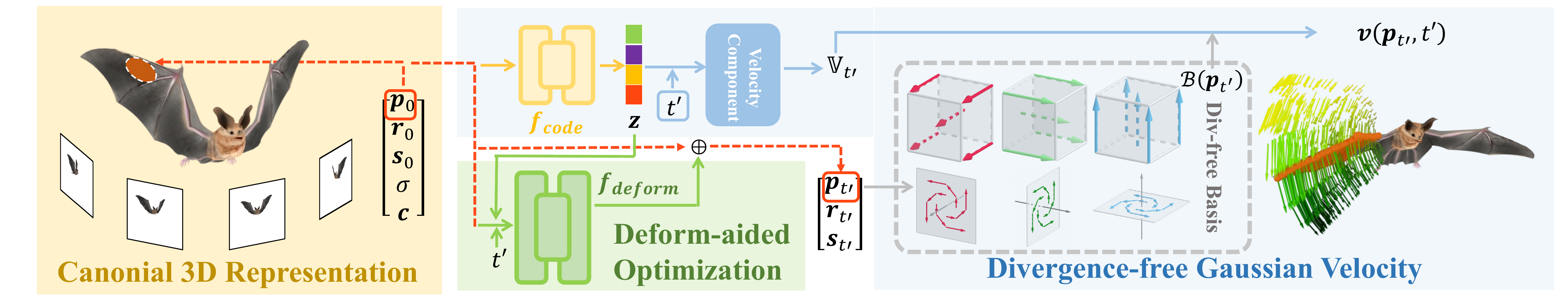}
%\vskip -0.15in
\caption{The leftmost block shows the canonical 3D Gaussians module, the middle bottom block shows the deformation-aided optimization module and the right block shows our divergence-free Gaussian velocity module.}
\label{fig:network}
\vskip -0.1in
\end{figure*}

\textbf{3D Physics Learning}: Inspired by recent physics-informed neural networks (PINN) \cite{Raissi2019}, a series of subsequent methods \cite{Raissi2020,Baieri2023,Wang2024,Im2024} learn various physical properties of 3D objects and scenes by converting different types of PDEs into loss functions as soft constraints. While achieving encouraging results in physics learning and future prediction, they often struggle to model complex physics in boundary regions. Another class of methods directly incorporates physics systems \cite{Xue2023,Jonathan2020,Mani2023,Zhong2024,Zhang2024,Liu2024a} such as graphs and springs into the network to learn the dynamics of specific targets including fluids and elastic objects. In our pipeline, we directly encode the generic divergence-free constraint into the learning process, thereby avoiding the use of PINN losses to train neural networks.  

\section{\nickname{}} \label{sec:method}
Given dynamic multi-view RGB videos with known camera poses and intrinsics over time, all frames are denoted as $\Big\{ \{\boldsymbol{I}_0^1 \cdots \boldsymbol{I}_0^n \cdots \boldsymbol{I}_0^N\} \cdots \{\boldsymbol{I}_T^1 \cdots \boldsymbol{I}_T^n \cdots \boldsymbol{I}_T^N\} \Big\}$, where  $N$ represents the total number of cameras and $T$ the greatest timestamp in training. 
Our pipeline is designed to simultaneously learn 3D geometry, appearance, and velocity information of the captured dynamic scene. As shown in Figure \ref{fig:network}, the first component - canonical 3D representation module - aims to learn a set of 3D Gaussian kernels to represent the 3D scene geometry and appearance in a canonical space, which is elaborated in Section \ref{sec:can_3d_rep}. The second component - neural divergence-free Gaussian velocity module - attempts to learn the 3D velocity of each Gaussian kernel, which is detailed in Section \ref{sec:neural_gau_vel}, while the third module - deformation-aided optimization - is discussed in Section \ref{sec:deform_opt}. 

\subsection{Canonical 3D Representation Module}\label{sec:can_3d_rep}

Following the vanilla 3DGS \cite{Kerbl2023}, this module aims to learn a set of 3D Gaussian kernels $\boldsymbol{P}_0$ to represent the canonical scene geometry and appearance at time $t=0$. Each kernel is parameterized by a 3D position $\boldsymbol{p}_0$, scaling $\boldsymbol{s}_0$, covariance matrix converted from quaternion $\boldsymbol{r}_0$, opacity $\sigma$, and color $\boldsymbol{c}$ converted from spherical harmonics (SH). Following prior works of dynamic Gaussian \cite{Wu2024,Yang2024}, we assume the opacity $\sigma$ and color $\boldsymbol{c}$ of each Gaussian are not changed over time, but constantly associated and transported with the kernel. 

We exactly follow 3DGS to train canonical kernels. Having the $N$ images at time $t=0$, we first initialize all canonical kernels $\boldsymbol{P}_0$ randomly or based on sparse point clouds created by SfM \cite{Schonberger2016}, and then project kernels into camera space, rendering 2D images and optimizing Gaussian parameters via $\ell_1$ and $\ell_{ssim}$ losses as used in 3DGS: 
\begin{equation}
\setlength{\abovedisplayskip}{3pt}
\setlength{\belowdisplayskip}{3pt}
    \underbrace{ \bigl\{\cdot\cdot (\boldsymbol{p}_0, \boldsymbol{r}_0, \boldsymbol{s}_0, \sigma, \boldsymbol{c}) \cdot\cdot \bigr\} }_{\boldsymbol{P}_0}  
    \mathrel{\mathop{\myrightleftarrows{\rule{1.cm}{0cm}}}^{\mathrm{render}}_{\mathrm{\ell_1 + \ell_{ssim}}}}
    \bigl\{\boldsymbol{I}^1_0 \cdot\cdot \boldsymbol{I}^n_0 \cdot\cdot \boldsymbol{I}^N_0 \bigr\}
\end{equation}

\subsection{Divergence-free Gaussian Velocity Module}\label{sec:neural_gau_vel}

This module aims to model the velocity of each Gaussian kernel. A na\"ive solution is to implement a simple network similar to the prior work NVFi \cite{Li2023c} to take each Gaussian as input and directly regress the corresponding 3D velocity, followed by commonly-used PINN losses to drive the estimated velocity to be physically meaningful, \textit{i.e.}, divergence-free. However, there are two key limitations. First, it is extremely hard to learn distinct 3D velocities for neighboring Gaussians exhibiting radically different motions, as a neural network just models a continuous and smooth function. Second, PINN losses are inefficient in learning physics priors as they involve dense sampling in both spatial and temporal dimensions during training. To this end, we tackle the problem of Gaussian velocity learning by introducing the following new technique ingredients. 

\textbf{Physics Code}: For each canonical Gaussian, we propose to learn a vector $\boldsymbol{z}$, which is regarded as a latent physics code to generally describe the underlying physical information possibly including mass, forces, \textit{etc.}. Basically, this code aims to explain the motion type of a Gaussian kernel over a whole period of time. Therefore, the code $\boldsymbol{z}$ is shared across all timestamps. In particular, we feed the position of each canonical Gaussian into an MLP-based network $f_{code}$, directly predicting its physics code $\boldsymbol{z}$:
\begin{equation}
\setlength{\abovedisplayskip}{3pt}
\setlength{\belowdisplayskip}{3pt}
\boldsymbol{z} = f_{code}(\boldsymbol{p}_0), \quad \boldsymbol{z}\in \mathcal{R}^L
\label{eq:physics_code}
\end{equation}
Technically, we may assign a learnable code to each canonical Gaussian independently. Nevertheless, such a strategy, albeit flexible, involves too many parameters to learn and we empirically find it is inferior as verified in our ablation study. Now, all canonical Gaussians $\boldsymbol{P}_0$ are as follows: 
\begin{equation}
\setlength{\abovedisplayskip}{3pt}
\setlength{\belowdisplayskip}{3pt}
 \boldsymbol{P}_0 \leftarrow \bigl\{\cdots (\boldsymbol{p}_0, \boldsymbol{r}_0, \boldsymbol{s}_0, \sigma, \boldsymbol{c}, \boldsymbol{z}) \cdots \bigr\} 
\end{equation}

\textbf{Divergence-free Velocity}: Having the physics code of a Gaussian kernel, our goal is to estimate its velocity field over the whole time period, without relying on the inefficient PINN losses. However, the estimated velocity field for each Gaussian must satisfy the basic divergence-free property. Otherwise, the extrapolated future frames will not be physically meaningful. 

To this end, we treat each canonical Gaussian  $\boldsymbol{p}_0$ as a rigid particle with its size and orientation, whose future motion is governed by a divergence-free velocity field $\boldsymbol{v}(\boldsymbol{p}_t, t)$, where $\boldsymbol{p}_t$ indicates the Gaussian position at time $t$. To learn the divergence-free velocity field $\boldsymbol{v}(\boldsymbol{p}_t, t)$ for a rigid particle, we propose to disentangle $\boldsymbol{v}(\boldsymbol{p}_t, t)$ into six basic velocity components: $\mathbb{V}_t\leftarrow[v_t^x, v_t^y, v_t^z, w_t^z, w_t^y, w_t^x]$ and a basis matrix $\mathcal{B}(\boldsymbol{p}_t)$, where $v_t^x / v_t^y / v_t^z$ represent the linear velocities along $x/y/z$ axes respectively, and $w_t^z / w_t^y / w_t^x$ represent the angular velocities of the rigid particle around $z/y/x$ axes respectively. Mathematically, the velocity field $\boldsymbol{v}(\boldsymbol{p}_t, t)$ is decomposed as follows:
\begin{equation}\label{eq:vel_decomp}
\setlength{\abovedisplayskip}{3pt}
\setlength{\belowdisplayskip}{3pt}
 \boldsymbol{v}(\boldsymbol{p}_t, t) = \mathbb{V}_t \cdot \mathcal{B}(\boldsymbol{p}_t), \quad \mathbb{V}_t\in \mathcal{R}^{1\times 6}\quad \mathcal{B}(\boldsymbol{p}_t) \in \mathcal{R}^{6\times 3}
\end{equation}
where $\mathcal{B}(\boldsymbol{p}_t)$ is a set of basis vectors defined as follows:
\begin{equation}\label{eq:divfree_basis}
\setlength{\abovedisplayskip}{3pt}
\setlength{\belowdisplayskip}{3pt}
\mathcal{B}(\boldsymbol{p}_t) = 
\begin{bmatrix}
1  &0  &0  &-p_t^y  &p_t^z  &0 \\
0  &1  &0  &p_t^x  &0  &-p_t^z \\
0  &0  &1  &0  &-p_t^x &p_t^y \\
\end{bmatrix}^T
\end{equation}
where $p_t^x / p_t^y / p_t^z$ represent the values of $\boldsymbol{p}_t$ at $x/y/z$ axes. The left three columns correspond to the linear velocities $v_t^x / v_t^y / v_t^z$, whereas the right three columns correspond to the angular velocities $w_t^z / w_t^y / w_t^x$. Note that, to guarantee the divergence-free property of $\boldsymbol{v}(\boldsymbol{p}_t, t)$, we will design $\mathbb{V}_t$ to be irrelevant to $\boldsymbol{p}_t$.

\begin{figure}[ht]%\vspace{-0.4cm}
\setlength{\abovecaptionskip}{ 4. pt}
\setlength{\belowcaptionskip}{ -10 pt}
\centering
   \includegraphics[width=1.\linewidth]{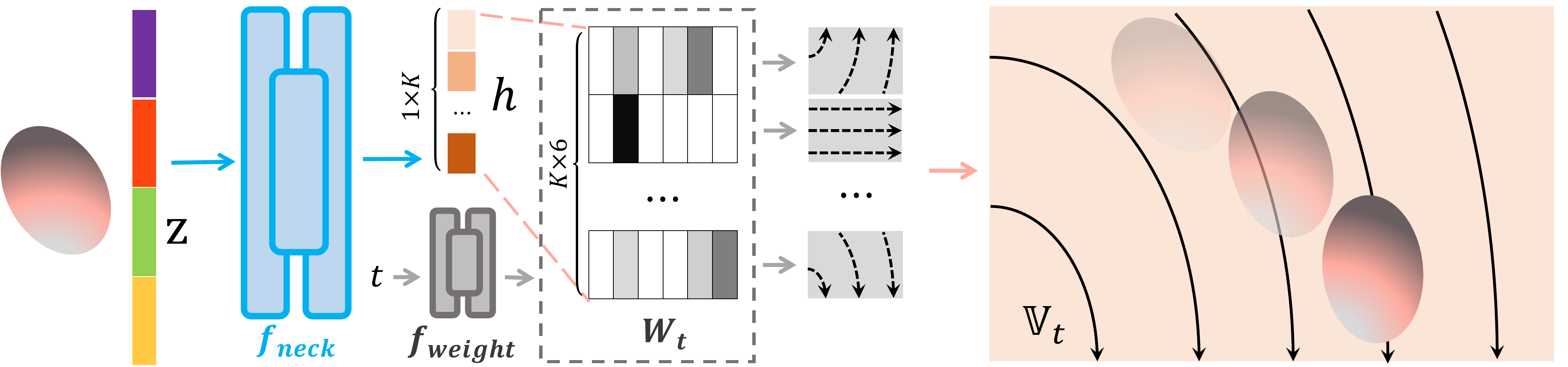}
\caption{An illustration of the network architecture to learn basic velocity component $\mathbb{V}_t$.}
\label{fig:basic_velo_comp}
\end{figure}

For any specific rigid particle at time $t$, we can easily collect its canonical position  $\boldsymbol{p}_0$ and physics code $\boldsymbol{z} = f_{code}(\boldsymbol{p}_0)$. To learn $\mathbb{V}_t$, a na\"ive solution is to feed $\boldsymbol{z}$ and $t$ into an MLP-based network, directly decoding a 6-dimensional vector. Nevertheless, we empirically find that this is too flexible and results in inferior performance as shown in our ablation study. To tackle this issue, we opt for a simple design illustrated in Figure \ref{fig:basic_velo_comp}.
%\vspace{-0.4cm}

In particular, we first feed the physics code $\boldsymbol{z}$ into several MLP layers $f_{neck}$, obtaining a bottleneck vector $\boldsymbol{h}\in \mathcal{R}^K$, where $K$ is a relatively small hyperparameter. 
In parallel, we feed the timestamp $t$ into another several MLP layers $f_{weight}$, obtaining a $K\times6$ vector. This is followed by a reshape operation and we get a matrix $\boldsymbol{W}_t\in \mathcal{R}^{K\times 6}$. The basic velocity component $\mathbb{V}_t$ is obtained by a matrix multiplication as follows:
\begin{equation}\label{eq:vel_components}
\setlength{\abovedisplayskip}{3pt}
\setlength{\belowdisplayskip}{3pt}
\mathbb{V}_t = \boldsymbol{h}\cdot \boldsymbol{W}_t = f_{neck}(\boldsymbol{z})\cdot f_{weight}(t)
\end{equation}
Basically, our insight behind this simple design is that: the physics code $\boldsymbol{z}$ tends to be decoded into $K$ types of motion patterns by $f_{neck}$, whereas $f_{weight}$ tends to generate a set of weights to select the decoded motion patterns at a particular timestamp $t$. Empirically, such a simple architecture achieves superior results and we leave more sophisticated designs for future exploration. All implementation details of the physics code, the velocity module, and the proof of divergence-free property are in Appendix \ref{app:proof}.

\subsection{Deformation-aided Optimization}\label{sec:deform_opt}

Regarding each Gaussian kernel, we need to effectively optimize its physics code $\boldsymbol{z}$ and basic velocity components $\mathbb{V}_t$, such that the corresponding per-Gaussian velocity field $\boldsymbol{v}(\boldsymbol{p}_t, t)$ is physically meaningful. However, this is particularly challenging because there are only multi-view images at sparse timestamps as supervision without any other priors. We empirically find that directly training all networks is hard to converge as shown in our ablation study. 

In this work, we introduce an auxiliary deformation field to aid the optimization process of our pipeline. In particular,  
we feed each canonical Gaussian kernel $\boldsymbol{p}_0$, timestamp $t$, and its physics code $\boldsymbol{z}$ into an MLP-based network $f_{deform}$, predicting the corresponding position displacement $\delta \boldsymbol{p}$, and the change of quaternion $\delta \boldsymbol{r}$ and scaling $\delta \boldsymbol{s}$ from timestamp 0 to $t$. Naturally, all Gaussians $\boldsymbol{P}_t$ at time $t$ can be computed by transporting all canonical Gaussians as follows. The deformation operations $\circ$ and $\odot$ follow \cite{Yang2024}. \vspace{-0.2cm}
\begin{align}\label{eq:deform}
\setlength{\abovedisplayskip}{-40pt}
\setlength{\belowdisplayskip}{-40pt}
(\delta \boldsymbol{p}, \delta\boldsymbol{r}, \delta\boldsymbol{s}) = f_{deform}(\boldsymbol{p}_0, t, \boldsymbol{z}), \quad\boldsymbol{z} = f_{code}(\boldsymbol{p}_0)\\ \nonumber
\boldsymbol{p}_t = \boldsymbol{p}_0+\delta\boldsymbol{p}, \quad \boldsymbol{r}_t = \boldsymbol{r}_0\circ\delta\boldsymbol{r}, \quad 
\boldsymbol{s}_t = \boldsymbol{s}_0\odot\delta\boldsymbol{s}\\ \nonumber
\boldsymbol{P}_t \leftarrow \bigl\{\cdots (\boldsymbol{p}_t, \boldsymbol{r}_t, \boldsymbol{s}_t, \sigma, \boldsymbol{c}, \boldsymbol{z}) \cdots \bigr\} 
\end{align} \vspace{-0.6cm}

With our design of physics code $\boldsymbol{z}$, basic velocity components $\mathbb{V}_t$, and the deformation field $f_{defom}$, we now discuss how to link and train them together, ultimately driving the learned per-Gaussian velocity field to be truly learned.

\textbf{First}, given training images $\big\{ \{\boldsymbol{I}_0^1 \cdot\cdot \boldsymbol{I}_0^N\} \cdot\cdot \{\boldsymbol{I}_T^1 \cdot\cdot \boldsymbol{I}_T^N\} \big\}$ and the canonical Gaussians $\boldsymbol{P}_0$, we sample a timestamp $t$ from training set and another timestamp $t'$, where $\Delta t = t - t'$ is predefined to be small enough. We can obtain Gaussians $\boldsymbol{P}_{t'}$ via the auxiliary deformation field $f_{deform}$(Eq \ref{eq:deform}):
\begin{equation}
\setlength{\abovedisplayskip}{3pt}
\setlength{\belowdisplayskip}{3pt}
\boldsymbol{P}_{t'} \leftarrow \bigl\{\cdots (\boldsymbol{p}_{t'}, \boldsymbol{r}_{t'}, \boldsymbol{s}_{t'}, \sigma, \boldsymbol{c}, \boldsymbol{z}) \cdots \bigr\} 
\end{equation}

\textbf{Second}, we feed the set of Gaussians $\boldsymbol{P}_{t'}$ at time $t'$ and the sampled interval $\Delta t$ into our interleaved mid-point Algorithm \ref{alg:vel_transport}, predicting the set of Gaussians $\boldsymbol{P}_t$ at time $t$ transported by our per-Gaussian velocity field.  \vspace{-0.4cm}
\begin{algorithm}[H]
\caption{ {\small Interleaved mid-point method.}
}
\label{alg:vel_transport}
\begin{algorithmic} 
\footnotesize
\State{\textbf{Input:}} 
(1) Gaussians at $t'$: $\bigl\{\cdot\cdot (\boldsymbol{p}_{t'}, \boldsymbol{r}_{t'}, \boldsymbol{s}_{t'}, \sigma, \boldsymbol{c}, \boldsymbol{z}) \cdot\cdot \bigr\}$, and (2) $\Delta t$ 
    
\State{\textbf{Output:}} 
Gaussians at $t$, \textit{i.e.,} at $(t' + \Delta t)$: $\bigl\{\cdot\cdot (\boldsymbol{p}_{t}, \boldsymbol{r}_{t}, \boldsymbol{s}_{t}, \sigma, \boldsymbol{c}, \boldsymbol{z}) \cdot\cdot \bigr\}$

\State{\phantom{xx}$\bullet\  \boldsymbol{v}(\boldsymbol{p}_{t'}, t) \leftarrow f_{neck}(\boldsymbol{z})\cdot f_{weight}(t')\cdot \mathcal{B}(\boldsymbol{p}_{t'})$, (ref to Eqs \ref{eq:vel_decomp}/\ref{eq:divfree_basis}/\ref{eq:vel_components})}

\State{\phantom{xx}$\bullet\ \boldsymbol{p}_{mid} \leftarrow \boldsymbol{p}_{t'} + \frac{\Dt}{2}\boldsymbol{v}(\boldsymbol{p}_{t'},t)$}

\State{\phantom{xx}$\bullet\  \boldsymbol{v}_{mid} \leftarrow f_{neck}(\boldsymbol{z})\cdot f_{weight}(t'+\frac{\Dt}{2})\cdot \mathcal{B}(\boldsymbol{p}_{mid})$}

\State{\phantom{xx}$\bullet\ \boldsymbol{p}_t \leftarrow \boldsymbol{p}_{t'} + \Dt \boldsymbol{v}_{mid}$}

\State{\phantom{xx}$\bullet$ \text{Convert quaternion to rotation:} $\mathbf{R}_{t'} \leftarrow \boldsymbol{r}_{t'}$}

\State{\phantom{xx}$\bullet\ \mathbf{R}_{t} \leftarrow (\mathbf{I}+\Dt \frac{\partial \boldsymbol{v}_{mid}}{\partial \boldsymbol{p}_{mid}})\mathbf{R}_{t'}$, (following the approximation in \cite{xie2023physgaussian}) }

\State{\phantom{xx}$\bullet$ \text{Convert rotation to quaternion:} $\boldsymbol{r}_{t} \leftarrow \mathbf{R}_{t} $ } 

\State{\phantom{xx}$\bullet$ \text{Assign $\boldsymbol{s}_{t'}$ to $\boldsymbol{s}_t$: $\boldsymbol{s}_t \leftarrow \boldsymbol{s}_{t'}$}, (each Gaussian is a rigid particle with its size fixed over time, ensuring the divergence-free property) } 

\\ \textbf{Return:} All gaussians at $t$: $\bigl\{\cdot\cdot (\boldsymbol{p}_{t}, \boldsymbol{r}_{t}, \boldsymbol{s}_{t}, \sigma, \boldsymbol{c}, \boldsymbol{z}) \cdot\cdot \bigr\}$
\end{algorithmic}
\end{algorithm}\vspace{-0.4cm}
\textbf{Third}, having the set of Gaussians $\boldsymbol{P}_t$ transported from $t'$ by our per-Gaussian velocity field, we project them to 2D space, rendering RGBs supervised by the total $N$ training views $\{\boldsymbol{I}_t^1 \cdot\cdot \boldsymbol{I}_t^N\}$ at time $t$ via the standard 3DGS losses:

\begin{equation}\label{eq:losses}
\setlength{\abovedisplayskip}{3pt}
\setlength{\belowdisplayskip}{3pt}
\{\boldsymbol{P}_0, f_{code}, f_{neck}, f_{weight}, f_{deform}\} \leftarrow (\ell_1 + \ell_{ssim})
\end{equation}
Implementation and training details are in Appendix \ref{app:implementation_details} \& \ref{app:optimization}.

\section{Experiments}
\label{sec:experiments}

\begin{table*}[t]\tabcolsep=0.2cm 
\centering
\caption{Quantitative results of all methods for both novel view interpolation and future frame extrapolation on Dynamic Object Dataset and Dynamic Indoor Scene Dataset, and future frame extrapolation on ParticleNeRF Dataset. Bold numbers indicate the best performance.}\vspace{-0.2cm}
\resizebox{1\linewidth}{!}{
\begin{tabular}{r|cccccc|cccccc|ccc}
\hline
\multirow{3}{*}{} & \multicolumn{6}{c|}{Dynamic Object Dataset} & \multicolumn{6}{c|}{Dynamic Indoor Scene Dataset} & \multicolumn{3}{c}{ParticleNeRF Dataset} \\ \cline{2-16} 
 & \multicolumn{3}{c|}{Interpolation} & \multicolumn{3}{c|}{Extrapolation} & \multicolumn{3}{c|}{Interpolation} & \multicolumn{3}{c|}{Extrapolation} & \multicolumn{3}{c}{Extrapolation} \\ \cline{2-16} 
 & PSNR$\uparrow$ & SSIM$\uparrow$ & \multicolumn{1}{c|}{LPIPS$\downarrow$} & PSNR$\uparrow$ & SSIM$\uparrow$ & LPIPS$\downarrow$ & PSNR$\uparrow$ & SSIM$\uparrow$ & \multicolumn{1}{c|}{LPIPS$\downarrow$} & PSNR$\uparrow$ & SSIM$\uparrow$ & LPIPS$\downarrow$ & PSNR$\uparrow$ & SSIM$\uparrow$ & LPIPS$\downarrow$ \\ 
 \hline
T-NeRF \citep{Pumarola2021} & 13.163 & 0.709 & \multicolumn{1}{c|}{0.353} & 13.818 & 0.739 & 0.324 & 24.944 & 0.742 & \multicolumn{1}{c|}{0.336} & 22.242 & 0.700 & 0.363 & - & - & - \\
D-NeRF \citep{Pumarola2021} & 14.158 & 0.697 & \multicolumn{1}{c|}{0.352} & 14.660 & 0.737 & 0.312 & 25.380 & 0.766 & \multicolumn{1}{c|}{0.300} & 20.791 & 0.692 & 0.349 & - & - & - \\
T-NeRF$_{PINN}$ & 15.286 & 0.794 & \multicolumn{1}{c|}{0.293} & 16.189 & 0.835 & 0.230 & 16.250 & 0.441 & \multicolumn{1}{c|}{0.638} & 17.290 & 0.477 & 0.618 & - & - & - \\
HexPlane$_{PINN}$ & 27.042 & 0.958 & \multicolumn{1}{c|}{0.057} & 21.419 & 0.946 & 0.067 & 25.215 & 0.763 & \multicolumn{1}{c|}{0.389} & 23.091 & 0.742 & 0.401 & - & - & - \\
NSFF \citep{Li2021c} & - & - & \multicolumn{1}{c|}{-} & - & - & - & 29.365 & 0.829 & \multicolumn{1}{c|}{0.278} & 24.163 & 0.795 & 0.289 & - & - & - \\
TiNeuVox \citep{Fang2022} & 27.988 & 0.960 & \multicolumn{1}{c|}{0.063} & 19.612 & 0.940 & 0.073 & 29.982 & 0.864 & \multicolumn{1}{c|}{0.213} & 21.029 & 0.770 & 0.281 & 20.407 & 0.888 & 0.102 \\
NVFi \citep{Li2023c}  & 29.027 & 0.970 & \multicolumn{1}{c|}{0.039} & 27.594 & 0.972 & 0.036 & \underline{30.675} & 0.877 & \multicolumn{1}{c|}{0.211} & 29.745 & 0.876 & 0.204 & 18.173 & 0.867 & 0.119 \\
DefGS \citep{Yang2024} & \underline{37.865} & 0.\underline{994} & \multicolumn{1}{c|}{\underline{0.007}} & 19.849 & 0.949 & 0.045 & 29.926 & \underline{0.916} & \multicolumn{1}{c|}{\underline{0.130}} & 21.380 & 0.819 & 0.188 & 19.205 & 0.900 & 0.083 \\ 
DefGS$_{nvfi}$ & 37.316 & \underline{0.994} & \multicolumn{1}{c|}{0.008} & \underline{28.749} & \underline{0.984} & \underline{0.013} & 30.170 & 0.915 & \multicolumn{1}{c|}{0.133} & \underline{31.096} & \underline{0.945} & \underline{0.077} & \underline{22.730} & \underline{0.931} & \underline{0.050}  \\ [+0.1em] 
\hline \textbf{\nickname{} (Ours)} & \textbf{39.393} & \textbf{0.995} & \multicolumn{1}{c|}{\textbf{0.005}} & \textbf{31.987} & \textbf{0.990} & \textbf{0.007} & \textbf{32.287} & \textbf{0.930} & \multicolumn{1}{c|}{\textbf{0.092}} & \textbf{35.019} & \textbf{0.966} & \textbf{0.051} & \textbf{26.657} & \textbf{0.956} & \textbf{0.030} \\[+0.1em] 
\hline
\end{tabular}
}
\label{tab:exp_extrapolation_nvfi}
\vspace{-0.2cm}
\end{table*}

\begin{table}\tabcolsep=0.1cm 
\caption{Quantitative results for both novel view interpolation and future frame extrapolation on \nickname{}-GoPro Dataset.}\vspace{-0.2cm}
\label{tab:exp_extrapolation_gopro}
% \footnotesize
\resizebox{1\linewidth}{!}{
\begin{tabular}{r|cccccc}
\hline
\multirow{3}{*}{} & \multicolumn{6}{c}{\nickname{}-GoPro Dataset}\\ \cline{2-7} 
 & \multicolumn{3}{c|}{Interpolation} & \multicolumn{3}{c}{Extrapolation}  \\ \cline{2-7} 
 & PSNR$\uparrow$ & SSIM$\uparrow$ & \multicolumn{1}{c|}{LPIPS$\downarrow$} & PSNR$\uparrow$ & SSIM$\uparrow$ & LPIPS$\downarrow$ \\ 
 \hline
TiNeuVox \citep{Fang2022} & 19.026 & 0.740 & \multicolumn{1}{c|}{0.319} & 20.600 & 0.760 & 0.292 \\
NVFi \citep{Li2023c}  & 18.947 & 0.739 & \multicolumn{1}{c|}{0.322} & 22.747 & 0.756 & 0.393 \\
DefGS \citep{Yang2024} & \underline{28.408} & \textbf{0.920} & \multicolumn{1}{c|}{\textbf{0.091}} & 21.344 & 0.858 & 0.170 \\ 
DefGS$_{nvfi}$ & 28.217 & \textbf{0.920} & \multicolumn{1}{c|}{\underline{0.092}} & \underline{26.431} & \underline{0.903} & \underline{0.124} \\ [+0.1em] 
\hline \textbf{\nickname{} (Ours)} & \textbf{28.451} & \textbf{0.920} & \multicolumn{1}{c|}{\textbf{0.091}} & \textbf{28.094} & \textbf{0.914} & \textbf{0.112} \\[+0.1em] 
\hline
\end{tabular}
}
\vspace{-0.4cm}
\end{table}

\phantom{Xx}\textbf{Datasets}: Our method focuses on learning physical information from dynamic videos for accurate future frame predictions, rather than fitting training views for interpolation. In this regard, the closest work to us is NVFi \cite{Li2023c}, and we mainly evaluate our method on its two synthetic dynamic datasets: \textbf{1) Dynamic Object Dataset} which consists of 6 dynamic objects undergoing different rigid or deformable motions; and \textbf{2) Dynamic Indoor Scene Dataset} which comprises 4 complex indoor scenes with 3 to 6 objects exhibiting challenging motions. For the sake of motion diversity, we also evaluate another synthetic \textbf{3) ParticleNeRF Dataset}\cite{abou2022particlenerf}, which includes various motions from robot manipulation to harmonic oscillation. All datasets have test splits for \textit{novel view interpolation} and \textit{future frame extrapolation}. More details of datasets are in Appendix \ref{app:dataset}.

While NVFi \cite{Li2023c} is evaluated on a real-world dataset, it just has two simple scenes selected from NVIDIA Dynamic Scene Dataset \citep{Yoon2020} which is not primarily collected for physics learning. Since this field of study is still in its infancy and there is lack of real-world datasets in high-quality, we opt to collect a new real-world dataset particularly for benchmarking 3D physics learning, which is the first of its kind to the best of our knowledge and will be made public. 

\begin{figure}[ht]
\vspace{-0.2cm}
\setlength{\abovecaptionskip}{ 2. pt}
\setlength{\belowcaptionskip}{ -8 pt}
\centering
\includegraphics[width=1.0\linewidth]{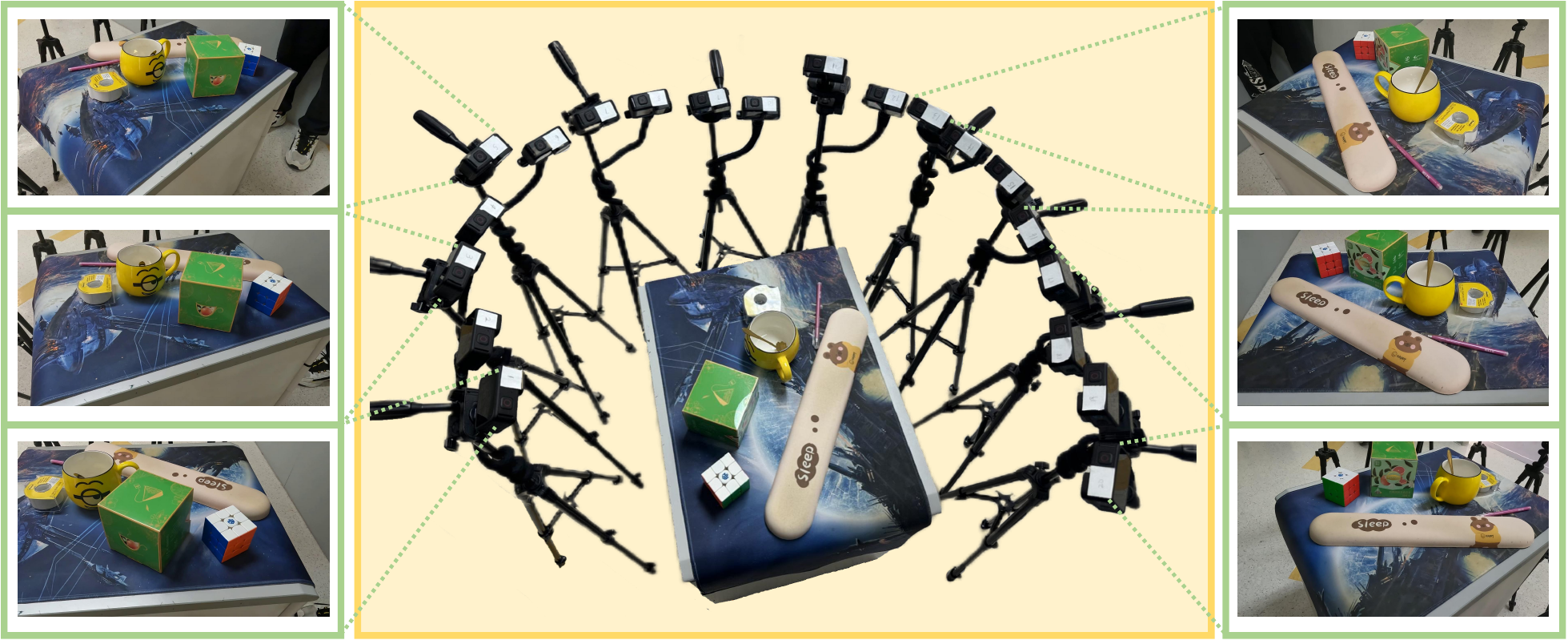}
\caption{The camera settings for our dataset collection.}
\label{fig:cam_setting}
\vspace{-0.1cm}
\end{figure}

\textbf{4) \nickname{}-GoPro Dataset}: We capture 6 dynamic scenes with 20 GoPro cameras (GoPro Hero 10 Black). All cameras are rigged on tripods and placed around an operation platform, as illustrated in Figure \ref{fig:cam_setting}. The 6 dynamic scenes consist of particularly challenging physical motions such as a moving object is tightly beside a static object, or a thin object is moving through a tiny hole of another object. 

We use voice to synchronize all videos and use COLMAP \cite{Schonberger2016} on the first frames of all 20 views to estimate camera poses. Since all cameras are fixed under a stabilizing mode, the camera poses do not change over time. For each dynamic scene, we select 89 frames from each view, and resize images to be a resolution of $960\times540$. We reserve the first 67 frames at 17 picked viewing angles as the training split, \textit{i.e.}, 1139 frames, while leaving the 67 frames at the remaining 3 viewing angles for evaluating \textit{novel view interpolation} within the training time period, \textit{i.e.}, 201 frames. We keep the last 22 frames at all 20 viewing angles for evaluating \textit{future frame extrapolation}, \textit{i.e.}, 440 frames in total. More details are in Appendix \ref{app:dataset}. 

\textbf{Baselines:} We choose the following baselines: \textbf{1) NVFi} \citep{Li2023c}. It is the closest work to us, but relies on PINN losses to learn physics priors whereas we directly build a divergence-free velocity field for each rigid particle. \textbf{2) T-NeRF} \citep{Pumarola2021}, \textbf{3) D-NeRF} \citep{Pumarola2021}, \textbf{4) NSFF} \citep{Li2021c}, and \textbf{5) TiNeuVox} \citep{Fang2022}. All these baselines adopt NeRF framework as the backbone, thus being short in 3D reconstruction compared to 3DGS. As our method adopts 3DGS, for a fair comparison, we also include the following two baselines. \textbf{6) DefGS} \citep{Yang2024}. This recent method Deformable 3D Gaussians is excellent to model dynamic 3D scenes for novel view interpolation, but lacks the ability for future prediction. In this regard, we build the most comparable baseline to us  \textbf{7) DefGS$_{nvfi}$} by combining DefGS with the velocity field proposed by NVFi. It shows strong ability in novel view synthesis (inherited from DefGS) and extrapolation ability (inherited from NVFi framework). To be fair,   
we combine the original deformation field of DefGS with exactly the same velocity field of NVFi. The same deformation-aided optimization as our \nickname{} is used to train DefGS$_{nvfi}$.
 
\textbf{Metrics:} The standard metrics \textbf{PSNR}, \textbf{SSIM}, and \textbf{LPIPS} are reported for RGB view synthesis in two tasks: novel view interpolation and future frame extrapolation.

\subsection{Evaluation for Future Frame Extrapolation}

We evaluate our method on all four datasets for both novel view interpolation and future frame extrapolation. All baselines are evaluated on the Dynamic Object Dataset and Dynamic Indoor Scene Dataset. We select 4 representative baselines to compare on our new \nickname{}-GoPro Dataset, including \textbf{TiNeuVox}, \textbf{NVFi}, \textbf{DefGS} and \textbf{DefGS$_{nvfi}$}.

\textbf{Results \& Analysis}: From Tables \ref{tab:exp_extrapolation_nvfi}\&\ref{tab:exp_extrapolation_gopro}, it can be seen that:  1) Our \nickname{} achieves the best performance on all datasets, clearly surpassing all baselines by a large margin on the challenging task of future frame extrapolation, showing that our method can indeed learn the underlying physical information, thus predicting accurate future motions. 2) Our method also achieves better scores on the task of novel view interpolation, notably outperforming the strong baselines DefGS and DefGS$_{nvfi}$. We hypothesize that the effective incorporation of physics priors allows our method to fully utilize all Gaussian kernels for representing 3D geometry and appearance. 
As shown in Figure \ref{fig:qual_res}, our method makes promising predictions, especially in dynamic multi-object scenes, such as the pen-through-tape scene, where only our method succeeds while others all fail. More quantitative and qualitative results are in Appendix \ref{app:results_object} to \ref{app:qualitative_all}.

\subsection{Analysis of Physics Code and Motion Patterns}

\begin{figure*}[t]
\centering
\includegraphics[width=1.0\linewidth]{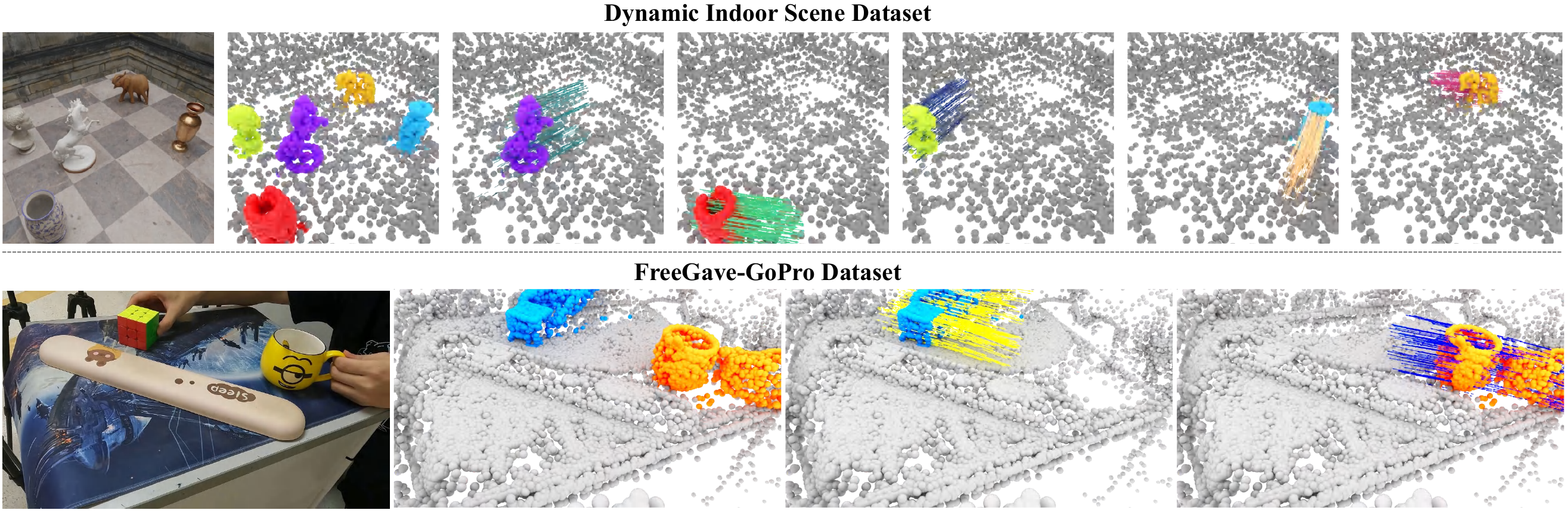}
%\vskip -0.15in
\caption{Qualitative results of decomposed motion patterns (colored objects) and their corresponding velocity fields (colored trajectories).}
\label{fig:seg}
\vspace{-0.3cm}
\end{figure*}

Our divergence-free Gaussian velocity module is designed to learn the velocity field for each Gaussian particle. With our design of bottleneck vector decomposition, our model can naturally group Gaussian kernels according to the learned motion patterns. To validate this nice property of bottleneck vector, we conduct the following experiments.
 
For each Gaussian, we query its learned velocity by uniformly sampling 10 timestamps to form a velocity trajectory $\{\vel_0, \ldots, \vel_{10}\}$. Then, we apply K-means \cite{macqueen1967knn} to group the learned bottleneck vectors $\boldsymbol{h} = f_{neck}(\z)$ into $C$ groups. In each group, we sample a number of Gaussian kernels and the queried velocity trajectories, followed by rendering to 2D space. We color the velocity trajectory based on the velocity direction. As shown in Figure \ref{fig:seg}, we can see that Gaussian kernels in the same group share the same velocity trajectory, while Gaussian kernels in different groups have totally different motion trajectories. 

We further quantitatively evaluate the objectness after grouping our learned physics codes on Dynamic Indoor Scene dataset. In particular, we follow \cite{gaussian_grouping} to render the group masks as 2D object segmentation masks for all 30 views over 60 timestamps on all 4 scenes, \textit{i.e.}, 7200 images in total. 
We compare our performance with \textbf{D-NeRF}, \textbf{NVFi}, \textbf{DefGS} and \textbf{DefGS$_{nvfi}$}. We follow NVFi to get the segmentation details of D-NeRF and NVFi. For the 3DGS based baselines, we also adopt OGC \cite{Song2022,Song2024b} to segment Gaussians based on scene flows calculated from their learned deformation fields. Implementation details are in Appendix \ref{app:segmentation}.  
In addition, we include a powerful image-based object segmentation method, Mask2Former \cite{Cheng2022} pre-trained on COCO \cite{Lin2014} dataset, as a fully-supervised baseline. 

As shown in Table \ref{tab:exp_decomposition} and Figure \ref{fig:qual_res}, we can see that our method achieves nearly perfect object segmentation results on all metrics, clearly outperforming all baselines. This shows that our learned physics codes truly represent object physical motion patterns which can be easily used to identify individual objects accordingly without needing any human labels to train a separate segmentation network.

\begin{table}\tabcolsep=0.1cm 
\centering
\footnotesize
\caption{Quantitative results of scene decomposition on the Synthetic Indoor Scene Dataset.}\vspace{-0.2cm}
\label{tab:exp_decomposition}
\begin{tabular}{rcccccc}
\toprule
            & AP$\uparrow$    & PQ$\uparrow$    & F1$\uparrow$    & Pre$\uparrow$    & Rec$\uparrow$    & mIoU$\uparrow$  \\ \midrule
Mask2Former \cite{Cheng2022} & 65.37 & 73.14 & 78.29 & \underline{94.83} & 68.88 & 64.42 \\
D-NeRF \cite{Pumarola2021} & 57.26 & 46.15 & 59.02 & 56.55 & 62.94 & 46.58 \\
NVFi \citep{Li2023c} & \underline{91.21} & \underline{78.74} & \underline{93.75} & 93.76 & \underline{93.74} & \underline{67.64} \\
DefGS \citep{Yang2024} & 51.73 & 57.60 & 66.43 & 63.21 & 70.07  & 54.46 \\
DefGS$_{nvfi}$ & 55.26 & 62.75 & 69.83 & 69.39 & 72.91 & 56.82  \\ \hline
\textbf{\nickname{} (Ours)} & \textbf{99.75} & \textbf{96.27} & \textbf{99.99} & \textbf{99.99} & \textbf{99.93} & \textbf{80.52} \\\bottomrule
\end{tabular}
\vspace{-0.4cm}
\end{table}

\subsection{Ablation Study}

Our framework comprises 3DGS as the scene representation backbone and the neural divergence-free Gaussian velocity module with a deformation-aided optimization module. We conduct the following groups of ablation experiments to validate the choices of our method. All ablation experiments are conducted on the Dynamic Objects dataset.

\textbf{(1) Replacing $f_{code}$ by a learnable code $\boldsymbol{z}$ in Eq \ref{eq:physics_code}}: Instead of feeding the position of each canonical Gaussian to $f_{code}$ to learn its physics code $\boldsymbol{z}$, we assign a learnable $\z \in \R^L$ for each Gaussian respectively. 

\textbf{(2) Removing divergence-free parametrization}: Instead of decomposing the velocity field by divergence-free parametrization in Eqs \ref{eq:vel_decomp}/\ref{eq:divfree_basis}/\ref{eq:vel_components}, we directly define the velocity field as $\vel(\p_t, t)=f_{neck}(\z)\cdot f_{motion}(\p_t, t)$, where $f_{motion}: \mathcal{R}^4\mapsto \mathcal{R}^K$ is an MLP-based network. Notably, now $f_{motion}$ is based on $\p_t$, so it is not divergence-free.

\textbf{(3) Removing bottleneck vector decomposition for velocity components in Eq \ref{eq:vel_components}}: Instead of decomposing velocity components network $\mathbb{V}_t$ by bottleneck vector in Eq \ref{eq:vel_components}, we directly define $\mathbb{V}_t$ as a single MLP: $\mathbb{V}_t=\mathrm{MLP}(\z, t)$.

\textbf{(4) Different choices of motion patterns $K$}: We compare 3 choices for the number of motion patterns in $f_{neck}$ / $f_{weight}$: $\{8, 16, 32\}$, while $K$=16 in our main experiments. 

\textbf{(5) Removing the auxiliary deformation field $f_{deform}$}: To demonstrate the necessity of our deformation-aided optimization scheme, we additionally train our velocity field without the deformation field. Particularly, we use Algorithm \ref{alg:vel_transport} to directly transport canonical Gaussians from $t=0$ to target time $t$. Since this requires autoregressive integration of all Gaussians by several steps, the memory cost accumulates rapidly, so we choose $\Dt = \frac{t}{2}$ to control the memory cost same as our main setting. 

\textbf{(6) Removing the physics code from deformation field $f_{deform}$}: Instead of feeding the physics code into the deformation field, we compare with the deformation field same as DefGS \cite{Yang2024}, \textit{i.e}, $(\delta \boldsymbol{p}, \delta\boldsymbol{r}, \delta\boldsymbol{s}) = f_{deform}(\boldsymbol{p}_0, t)$.

\textbf{(7) Removing the scaling deformation from deformation field}: We remove the scaling deformation from the output of $f_{deform}$, forcing the motion of each Gaussian to be volume-preserving all time, \textit{i.e.}, $(\delta \boldsymbol{p}, \delta\boldsymbol{r}) = f_{deform}(\boldsymbol{p}_0, t, \boldsymbol{z})$. Gaussian scales are not updated all time.

\setlength{\abovecaptionskip}{0 pt}
\begin{table} \tabcolsep=0.1cm 
\caption{Quantitative results of ablation studies on Dynamic Object Dataset. } \vspace{0.2cm}
\label{tab:exp_ablation_obj}
\footnotesize
\resizebox{1\linewidth}{!}{
\begin{tabular}{ccccc|ccc}
\hline
 &  &  &  &  & \multicolumn{3}{c}{Extrapolation} \\ \cline{6-8} 
  & Code $\z$ & $f_{deform}$ & $\vel(\p_t,t)$ & $K$ & PSNR$\uparrow$ & SSIM$\uparrow$ & LPIPS$\downarrow$ \\ \hline
(1) & learnable & full & full & 16 & 25.961 & 0.975 & 0.025 \\ \hline
(2) & field & full & w/o $\mathcal{B}(\p_t)$ & 16 & 29.400 & 0.986 & 0.010 \\
(3) & field & full & w/o decomp & 16 & 29.432 & 0.985 & \underline{0.009} \\
(4) & field & full & full & 8 & 30.972 & \underline{0.989} & \underline{0.009} \\
(4) & field & full & full & 32 & 31.438 & \textbf{0.990} & \textbf{0.007} \\ \hline
(5) & field & \N & full & 16 & 17.927 & 0.922 & 0.088 \\ 
(6) & field & w/o $\z$ & full & 16 & 31.217 & 0.988 & \underline{0.009} \\ 
(7) & field & w/o $\delta\s$ & full & 16 & \underline{31.704} & \textbf{0.990} & \textbf{0.007} \\ \hline
\textbf{\nickname{}} & field & full & full & 16 & \textbf{31.987} & \textbf{0.990} & \textbf{0.007} \\ \hline
\end{tabular}
}
\vspace{-0.4cm}
\end{table}

\textbf{Results \& Analysis}: From Table \ref{tab:exp_ablation_obj},  
we can see that: 1) Removing deformation field is the most influential. Since autoregressive integration is needed in this case, in order to fit the memory cost, we have to use a large integration time interval, as large as half of the target time. This incurs instability and noise in training, making the network totally unable to converge. The noise also leads to large gradients for all Gaussian kernels, resulting in unreasonable densification and ultimately making the network totally collapse on Dynamic Indoor Scene Dataset. 2) Replacing the physics code network by assigning a learnable physics code to each Gaussian kernel shows the next greatest impact on performance, demonstrating that the local continuity property of an MLP is helpful for network converging. 3) Either removing divergence-free parametrization or removing motion pattern decomposition diminishes the performance, due to the drastically increase in the degree of freedom and complexity of velocity field. 4) Feeding physics codes as input to the deformation field can enhance the performance. 5) Too few motion patterns decomposition hurts the performance, while over-parametrization is not sensitive. Complete ablation results are in Appendix \ref{app:ablation_all}.

\begin{figure*}[t]
\setlength{\abovecaptionskip}{ 4. pt}
\setlength{\belowcaptionskip}{ -10 pt}
\centering
   \includegraphics[width=0.92\linewidth]{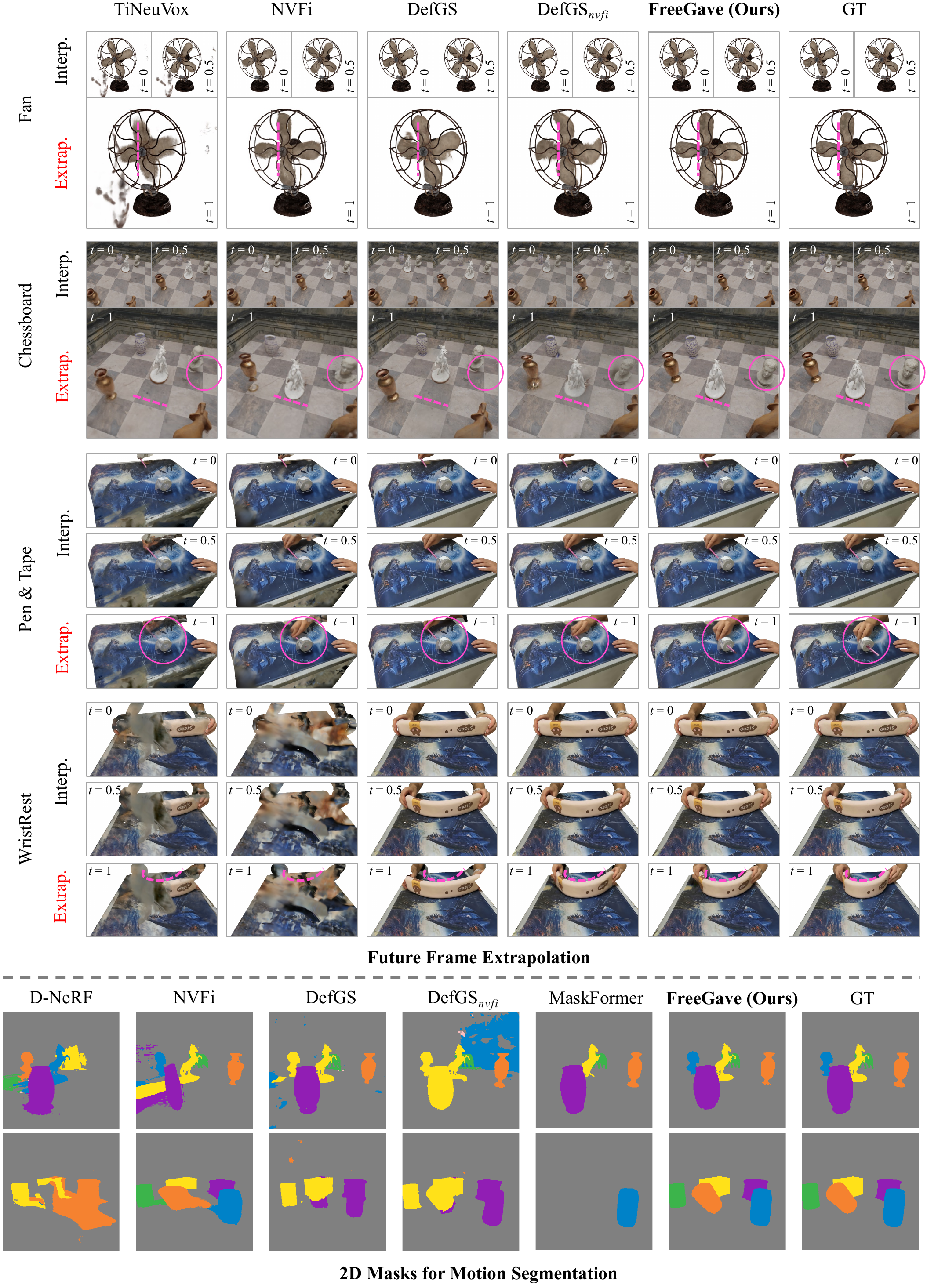}
\caption{Qualitative results of our method and baselines for future frame extrapolation and unsupervised motion segmentation.}
\label{fig:qual_res}
\end{figure*}

\vspace{-0.1cm}
\section{Conclusion}
\label{sec:con}

Our method extends 3D Gaussian splatting to simultaneously learn 3D scene geometry, appearance, and the critical physics information just from multi-view videos, without needing any object priors. By introducing a physics code and a divergence-free velocity field for each Gaussian kernel, our method can effectively learn precise physical motion patterns without needing the inefficient PINN losses. Extensive experiments on three public datasets and our newly captured challenging real-world dataset demonstrate extraordinary performance of our method for future frame prediction and object motion decomposition. It would be interesting to apply our method in 4D generation where 3D physics learning is crucial to generate plausible motions.

\phantom{xx}

\noindent\textbf{Acknowledgment:} 
This work was supported in part by Research Grants Council of Hong Kong under Grants 25207822 \& 15225522.

\clearpage
{
\small
\bibliographystyle{ieeenat_fullname}
\bibliography{references}
}

\appendix
\clearpage
\setcounter{page}{1}
\maketitlesupplementary

\section{Proof of Divergence-free Property}
\label{app:proof}
The velocity field is defined as:
\begin{equation}
    \vel(\p_t, t) = \V_t\cdot\mathcal{B}(\p_t)=\sum_{k=1}^{6}{\V_t^k\mathcal{B}^k(\p_t)}.
\end{equation}

In order to prove the divergence-free property, we just need to show $\nabla_{\p_t}\cdot\vel(\p_t,t)=0$. Since $\V_t$ is totally irrelevant to $\p_t$, we only need to show each basis vector function follows $\nabla_{\p_t}\cdot\mathcal{B}^k(\p_t)=0$, 
 \textit{i.e.}: 

\begin{align}
    \nabla_{\p_t}\cdot\vel(\p_t, t) & = \nabla_{\p_t}\cdot(\V_t\cdot\mathcal{B}(\p_t)) \\
    & =\sum_{k=1}^{6}{\V_t^k\nabla_{\p_t}\cdot\mathcal{B}^k(\p_t)}=0.
\end{align}

Next, we show each of the basis vector function is divergence-free:
\begin{align}
    \nabla_{\p_t}\cdot\mathcal{B}^1(\p_t) & =\nabla_{\p_t}\cdot\begin{bmatrix} 1 & 0 & 0 \end{bmatrix} = \frac{\partial 1}{\partial p_t^x} = 0; \\
    \nabla_{\p_t}\cdot\mathcal{B}^2(\p_t) & =\nabla_{\p_t}\cdot\begin{bmatrix} 0 & 1 & 0 \end{bmatrix} = \frac{\partial 1}{\partial p_t^y} = 0; \\ 
    \nabla_{\p_t}\cdot\mathcal{B}^3(\p_t) & =\nabla_{\p_t}\cdot\begin{bmatrix} 0 & 0 & 1 \end{bmatrix} = \frac{\partial 1}{\partial p_t^z} = 0; \\
    \nabla_{\p_t}\cdot\mathcal{B}^4(\p_t) & =\nabla_{\p_t}\cdot\begin{bmatrix} -p_t^y & p_t^x & 0 \end{bmatrix} \\
    & = \frac{\partial -p_t^y}{\partial p_t^x} + \frac{\partial p_t^x}{\partial p_t^y} = 0; \\
    \nabla_{\p_t}\cdot\mathcal{B}^5(\p_t) & =\nabla_{\p_t}\cdot\begin{bmatrix} p_t^z & 0 & -p_t^x \end{bmatrix} \\
    & = \frac{\partial p_t^z}{\partial p_t^x} + \frac{\partial -p_t^x}{\partial p_t^z} = 0; \\
    \nabla_{\p_t}\cdot\mathcal{B}^6(\p_t) & =\nabla_{\p_t}\cdot\begin{bmatrix} 0 & -p_t^z & p_t^y \end{bmatrix} \\
    & = \frac{\partial -p_t^z}{\partial p_t^y} + \frac{\partial p_t^y}{\partial p_t^z} = 0. 
\end{align}

\section{Implementation Details}
\label{app:implementation_details}
We implement our networks by MLPs, and the detailed configurations are as follows:
\begin{itemize}
    \item $f_{code}$: This network includes 4 MLP layers and each hidden layer has a size of 128 neurons followed by ReLU. In addition, we make a degree of 8 positional encoding for input positions. We set the dimension of the output physics code $L$ as 16.
    \item $f_{neck}$: This network has MLP layers as $L \mapsto 4L \mapsto 4L \mapsto K$. We choose the bottleneck vector dimension $K$ as 32 for Dynamic Indoor Scene dataset and 16 for other two datasets. 
    \item $f_{weight}$: This network includes 5 MLP layers and each hidden layer has a size of 128 neurons. In addition, we make a degree of 8 positional encoding for input positions, and we add a ResNet connection on the 3rd layer. The output vector has a dimension of $K*6$ and is reshaped as a matrix of $K \times 6$.
    \item $f_{deform}$: For two synthetic datasets, this network includes 6 MLP layers and each hidden layer has a size of 128 neurons, while it includes 8 MLP layers and each hidden layer has a size of 256 neurons for the challenging \nickname-GoPro dataset. In addition, we also make a degree of 8 positional encoding for input positions, and the physics code $\z$ is concatenated to the encoded position. We also add a ResNet connection on the third layer.
\end{itemize}

\section{Details of Deformation-aided Optimization}
\label{app:optimization}
We train our models on all datasets on a single NVIDIA 3090 GPU. 
We normalize the total time span in all datasets to be 1.
$\Dt$ is set to be 1/60 in both Dynamic Object dataset and Dynamic Indoor Scene dataset, while it is set as 1/88 in the challenging \nickname-GoPro dataset.

\section{Details of All Datasets}
\label{app:dataset}

\phantom{Xx}\textbf{Dynamic Object Dataset} \cite{Li2023c}: This dataset contains 6 moving objects with a white background, and the corresponding motions include: 1. part-wise rigid motions with accelerations, \textit{i.e.}, rotating fan, freely falling basketball in a gravitational field, and rotating telescope; 2. self-propelling deformable objects, \textit{i.e.}, a bat flapping wings, a swimming whale, and a swimming shark. 
Each scene contains 15 viewing angles, where the first 46 frames from 12 selected viewing angles are used as training split, \textit{i.e.}, 552 frames in total, and the first 46 frames from the other 3 viewing angles are used for evaluating novel view interpolation within the training time period, \textit{i.e.}, 138 frames in total. All the remaining 14 frames from 15 viewing angles are used to evaluate future frame extrapolation, \textit{i.e.}, 210 frames. 

\textbf{Dynamic Indoor Scene Dataset} \cite{Li2023c}: This dataset contains 4 indoor scenes, each containing 3 to 5 moving objects, and each moving object is undergoing different rigid motions. Each scene contains 30 viewing angles, where the first 46 frames from 25 selected viewing angles are used as the training split, \textit{i.e.}, 1150 frames in total, and the first 46 frames from the other 5 viewing angles are used for evaluating novel view interpolation within the training time period, \textit{i.e.}, 230 frames in total. All the remaining 14 frames from 30 viewing angles are used to evaluate future frame extrapolation, \textit{i.e.}, 450 frames. 

\textbf{ParticleNeRF Dataset} \cite{abou2022particlenerf}: This dataset includes 6 challenging dynamic objects. 
\begin{itemize}
    \item \textbf{Object \#1: Robot}. This scene includes a robot arm waving from one side to another side.
    \item \textbf{Object \#2: Robot Task}. This scene shows a robot arm putting a box onto a sliding platform.
    \item \textbf{Object \#3: Wheel}. This scene includes a constant rotating wheel.
    \item \textbf{Object \#4: Spring}. This scene shows a box tied on a spring, which undergoes a harmonic oscillation motion. The training and the test observation period in total form a whole oscillation period. 
    \item \textbf{Object \#5: Pendulums}. This scene includes two swing pendulums, each undergoing harmonic oscillation motion. The training and the test observation period in total form a whole oscillation period. 
    \item \textbf{Object \#6: Cloth}. This scene includes a flat cloth being folded.
\end{itemize}

Each scene contains 40 viewing angles. For Object \#1 \& \#2, we choose the first 53 frames from 36 selected viewing angles as the training split, \textit{i.e.} 1908 frames in total, and the first 53 frames from the other 4 viewing angles for evaluating novel view interpolation within the training time period, \textit{i.e.}, 212 frames in total. All the remaining 17 frames from 40 viewing angles are used to evaluate future frame extrapolation, \textit{i.e.}, 680 frames. 
For Object \#3 \& \#4 \& \#5 \& \#6, the first 104 frames from 36 selected viewing angles are used as the training split, \textit{i.e.}, 3744 frames in total, and the first 104 frames from the other 4 viewing angles are used for evaluating novel view interpolation within the training time period, \textit{i.e.}, 416 frames in total. All the remaining 36 frames from 40 viewing angles are used to evaluate future frame extrapolation, \textit{i.e.}, 1440 frames. 

\textbf{\nickname-GoPro Dataset}: This dataset includes 6 challenging real-world dynamic scenes. 
\begin{itemize}
    \item \textbf{Scenes \#1/\#2: Pen \& Tape}. There is a person holding a pen and trying to pass it through the hole of a static tape. The difference between these two scenes is that there are more static objects in the second scene, introducing more visual occlusions and requiring more accurate separation between moving areas and static areas. The difficulty lies in the motion of one object which is going to penetrate through another in future.  
    \item \textbf{Scene \#3: Box}. This scene contains a drawer-like box, and a person is trying to close it. The difficulty lies in a tight combination of the moving part and the static part of the box, especially in future. 
    \item \textbf{Scene \#4: Hammer}. This scene contains a hammer moving on the topside of a box. The difficulty lies in the direct contact of moving objects and static objects, which requires sharp separation of diverse motion patterns in order to keep right static/moving states in future. 
    \item \textbf{Scene \#5: Collision}. This scene contains a cube and a cup moving towards each other. The difficulty is the different directions of two motions. It is hard to keep the shapes of these two objects in future. 
    \item \textbf{Scene \#6: Wrist Rest}. A person is trying to bend a wrist rest. The difficulty is that the object is deformable and the motion is thus not rigid or part-wise rigid. 
\end{itemize}

\section{Quantitative Results on NVIDIA Dynamic Scene Dataset}
Though not primarily collected for physics learning, we also evaluate our model on two simple scenes from NVIDIA Dynamic Scene Dataset \citep{Yoon2020} selected by NVFi \cite{Li2023c}. The quantitative results are shown in Table \ref{tab:exp_nvidia}.

\begin{table}[!h]\tabcolsep=0.1cm 
\centering
\vspace{-0.3cm}
\caption{Results on NVIDIA Dynamic Scene Dataset.} %\vspace{-0.35cm}
\resizebox{1.0\linewidth}{!}{
\begin{tabular}{l|cccccc}
\hline
& \multicolumn{3}{c|}{Interpolation} & \multicolumn{3}{c}{Extrapolation} \\ \cline{2-7} 
& PSNR$\uparrow$ & SSIM$\uparrow$ & \multicolumn{1}{c|}{LPIPS$\downarrow$} & PSNR$\uparrow$ & SSIM$\uparrow$ & LPIPS$\downarrow$ \\ 
\hline
T-NeRF & 23.078 & 0.684 & \multicolumn{1}{c|}{0.355} & 21.120 & 0.707 & 0.358  \\
D-NeRF & 22.827 & 0.711 & \multicolumn{1}{c|}{0.309} & 20.633 & 0.709 & 0.327 \\
TiNeuVox & \textbf{28.304} & 0.868 & \multicolumn{1}{c|}{0.216} & 24.556 & 0.863 & 0.215 \\
T-NeRF$_{PINN}$ & 18.443 & 0.597 & \multicolumn{1}{c|}{0.439} & 17.975 & 0.605 & 0.428 \\
HexPlane$_{PINN}$ & 24.971 & 0.818 & \multicolumn{1}{c|}{0.281} & 24.473 & 0.818 & 0.279 \\
NVFi & 27.138 & 0.844 & \multicolumn{1}{c|}{0.231} & \underline{28.462} & 0.876 & 0.214 \\
DefGS & 26.662 & \underline{0.893} & \multicolumn{1}{c|}{\underline{0.127}} & 24.240 & 0.895 & 0.140  \\ 
DefGS$_{nvfi}$ & 26.972 & 0.890 & \multicolumn{1}{c|}{0.128} & 27.529 & \underline{0.927} & \underline{0.102}  \\ [+0.1em] 
\hline 
\textbf{\nickname{} (Ours)} & \underline{27.345} & \textbf{0.896} & \multicolumn{1}{c|}{\textbf{0.097}} & \textbf{29.005} & \textbf{0.933} & \textbf{0.072}  \\[+0.1em] 
\hline
\end{tabular}
}
\label{tab:exp_nvidia}
\vspace{-0.3cm}
\end{table}

\section{Quantitative Results on Collision Cases}

We evaluate on two more scenes with collisions: We extend the \textit{dining} scene of Dynamic Indoor Scene Dataset into two collision cases with different collision patterns. The first scene has 28 frames $\times$ 25 views for training without observing collision, 28 frames $\times$ 5 views for interpolation, and 8 $\times$ 30 views for extrapolation where the collision happens. The second scene has 46 frames $\times$ 25 views for training with the collision observed, 46 frames $\times$ 5 views for interpolation, and 14 $\times$ 30 views for extrapolation. 

\begin{table}[h]\tabcolsep=0.2cm 
\centering
\vspace{-0.3cm}
\caption{Results on four scenes of oscillations or collisions.} %\vspace{-0.4cm}
\resizebox{1\linewidth}{!}{
\begin{tabular}{l|ccc|ccc}
\hline
\multirow{3}{*}{} & \multicolumn{6}{c}{Collisions} \\ \cline{2-7} 
 & \multicolumn{3}{c|}{Interpolation} & \multicolumn{3}{c}{Extrapolation} \\ \cline{2-7} 
 & PSNR$\uparrow$ & SSIM$\uparrow$ & LPIPS$\downarrow$ & PSNR$\uparrow$ & SSIM$\uparrow$ & LPIPS$\downarrow$ \\ 
 \hline
TiNeuVox & 23.429 & 0.794 & 0.277 & 20.794 & 0.807 & 0.250 \\
NVFi  & 20.301 & 0.690 & 0.413 & 22.917 & 0.780 & 0.313 \\
DefGS & 29.411 & 0.894 & 0.117 & 23.129 & 0.867 & 0.122 \\ 
DefGS$_{nvfi}$ & 29.424 & 0.894 & 0.118 & 28.017 & 0.907 & 0.081 \\ [+0.1em] 
\hline \textbf{\nickname{} (Ours)} & \textbf{29.971} & \textbf{0.916} & \textbf{0.074} & \textbf{28.426} & \textbf{0.912} & \textbf{0.058} \\[+0.1em] 
\hline
\end{tabular}
}
\label{tab:exp_oscillation_collision}
\vspace{-0.4cm}
\end{table}

\section{Analysis of Computational Costs}

We calculate the average time and memory consumption in training, the average speed and memory consumption in test, and model sizes for most baselines in Table \ref{tab:cost}. 

We can see that: 1) our method has a clear advantage over the strong baseline DefGS$_{nvfi}$ in terms of time and memory cost in training, thanks to our new and more efficient divergence-free velocity module over the PINN loss; 2) our method is generally better or on par with other baselines (TiNeuVox / NVFi / DefGS) in computation cost of training and test, but our method demonstrates significantly better extrapolation results (as shown in Tables \ref{tab:exp_extrapolation_nvfi}\&\ref{tab:exp_extrapolation_gopro}).   

%\vspace{-0.25cm}
\begin{table*}[h]\tabcolsep=0.2cm 
\centering
\caption{The average time (hours) and GPU memory (GB) cost for training, the inference speed (fps), GPU memory (GB), and model size (MB) for test on all three datasets.}%\vspace{-0.2cm}
\resizebox{1\linewidth}{!}{
\begin{tabular}{l|cc|ccc|cc|ccc|cc|ccc}
\hline
 & \multicolumn{5}{c|}{Dynamic Object Dataset} & \multicolumn{5}{c|}{Dynamic Indoor Scene Dataset} & \multicolumn{5}{c}{\nickname-GoPro Dataset} \\ \cline{2-16} 
 & \multicolumn{2}{c|}{Training} & \multicolumn{3}{c|}{Test} & \multicolumn{2}{c|}{Training} & \multicolumn{3}{c|}{Test} & \multicolumn{2}{c|}{Training} & \multicolumn{3}{c}{Test} \\ \cline{2-16} 
 & Time$\downarrow$ & Mem$\downarrow$ & fps$\uparrow$ & Mem$\downarrow$ & Size$\downarrow$ & Time$\downarrow$ & Mem$\downarrow$ & fps$\uparrow$ & Mem$\downarrow$ & Size$\downarrow$ & Time$\downarrow$ & Mem$\downarrow$ & fps$\uparrow$ & Mem$\downarrow$ & Size$\downarrow$  \\ 
 \hline
TiNeuVox & \textbf{0.5} & 8.0 & 0.40 & 2.3 & \underline{49.8} & \textbf{0.6} & \underline{8.2} & 0.51 & \underline{3.7} & \textbf{49.9} & \textbf{0.6} & \underline{9.4} & 0.16 & \underline{4.7} & \textbf{49.6} \\
NVFi  & 2.2 & 22.6 & 0.11 & 16.5 & 114.7 & 2.3 & 21.5 & 0.34 & 16.8 & 107.9 & 2.3 & 23.3 & 0.03 & 23.1 & 121.4 \\
DefGS & \underline{0.8} & \underline{6.4} & \underline{19.40} & \underline{5.0} & 53.1 & \underline{0.8} & \textbf{5.9} & \textbf{21.9} & 4.0 & 98.1 & \underline{1.3} & \textbf{8.4} & \textbf{3.22} & \textbf{3.4} & \underline{88.5} \\ 
DefGS$_{nvfi}$ & 2.1 & 22.7 & 13.59 & 5.2 & 54.5 & 6.0 & 26.4 & 10.98 & 4.1 & 101.2 & 8.0 & 32.6 & 2.90 & 5.8 & 92.1 \\ 
\hline \textbf{\nickname{} (Ours)} & \underline{0.8} & \textbf{5.7} & \textbf{19.70} & \textbf{4.0} & \textbf{27.0} & 1.5 & 10.4 & \underline{13.60} & \textbf{3.1} & \underline{55.8} & 1.9 & 16.1 & \underline{3.03} & 8.1 & 115.7 \\[+0.1em] 
\hline
\end{tabular}
}
\label{tab:cost}
%\vspace{-0.25cm}
\end{table*}

\section{Limitation of Our Model}

The main limitation is that our method would fail to predict abrupt motions, such as an explosion, primarily because the underlying physics rules are unable to be observed or learned from visual frames. 

\section{Details of Segmenting Motion Patterns}
\label{app:segmentation}

For our models and the DefGS / DefGS$_{nvfi}$ baselines, we render the segmentation masks after grouping all learned Gaussian kernels. We follow the rendering module in Gaussian-Grouping \cite{gaussian_grouping} to obtain segmentation masks. Gaussian-Grouping renders hidden segmentation features in a size of 16. Therefore, we directly expand our one-hot object group into 16 channels and then render masks. More details are as follows. 

\subsection{More Details of Our \nickname{}}
\label{app:seg_ours}
We segment our well-trained Gaussian kernels by their bottleneck vectors $\boldsymbol{h}=f_{neck}(\z)$. To be specific, we build a grouping feature vector for each Gaussian kernel as $\boldsymbol{h}\oplus \lambda \p_0$, where $\oplus$ means concatenation and $\lambda$ is a hyperparameter, working as smoothing regularization. Then the Gaussian kernels are simply grouped by K-means algorithm with respect to this built features into C groups. 

We choose $\lambda$ as 0 for Genome House and Chessboard scene, and 0.5 for Factory and Dining Table scene. C is set as 13 for Dining Table scene and 8 for other three scenes.

\subsection{More Details of Segmenting DefGS / DefGS$_{nvfi}$}
\label{app:seg_ogc}

Given a well-trained DefGS or DefGS$_{nvfi}$ model with $N$ canonical Gaussian kernels, we first assign learnable per-Gaussian object codes $\boldsymbol{O} \in {(0, 1)}^{N \times K}$ to all Gaussian kernels, where $K$ is the maximum number of objects that is expected to appear in the scene.

After that, we query the position displacements for Gaussians kernels from the well-trained deformation field at time $0$ and $t$ respectively, thus obtaining the Gaussians $\boldsymbol{P}_0$ at time $0$ and $\boldsymbol{P}_t$ at time $t$. Then the per-Gaussian scene flows $\boldsymbol{M}_t$ from time $0$ to $t$ is calculated as $\boldsymbol{M}_t = \boldsymbol{P}_t - \boldsymbol{P}_0$.

Lastly, two losses proposed in OGC \cite{Song2022} are computed on the learnable object codes. \textbf{1) Dynamic rigid consistency:} For the $k^{th}$ object, we first retrieve its (soft) binary mask $\boldsymbol{O}^k$, and feed the tuple $\{\boldsymbol{P}_0, \boldsymbol{P}_t, \boldsymbol{O}^k\}$ into the weighted-Kabsch algorithm to estimate its transformation matrix $\boldsymbol{T}_k \in \mathbb{R}^{4\times 4}$ belonging to $SE(3)$ group. Then the dynamic loss is computed as:
\begin{equation*}
    \ell_{dynamic} = \frac{1}{N} \sum_{\boldsymbol{p} \in \boldsymbol{P}_0} \Big\| \Big(\sum_{k=1}^K {o}_{\boldsymbol{p}}^{k} \cdot (\boldsymbol{T}_k \circ \boldsymbol{p}) \Big) - (\boldsymbol{p} + \boldsymbol{m}_t) \Big\|_2
\end{equation*}
where ${o}_{\boldsymbol{p}}^{k}$ represents the probability of being assigned to the $k^{th}$ object for a specific point $\boldsymbol{p}$, and $\boldsymbol{m}_t \in \mathbb{R}^{3}$ represents the motion vector of $\boldsymbol{p}$ from time $0$ to $t$. The operation $\circ$ applies the rigid transformation to the point. This loss aims to discriminate objects with different motions. \textbf{2) Spatial smoothness:} For each point $\boldsymbol{p}$ in $\boldsymbol{P}_0$, we first search $H$ nearest neighboring points. Then the smoothness loss is defined as:
\begin{equation}
    \ell_{smooth} = \frac{1}{N} \sum_{\boldsymbol{p} \in \boldsymbol{P}_0} \Big( \frac{1}{H}\sum_{h=1}^H \| \boldsymbol{o}_{\boldsymbol{p}} - \boldsymbol{o}_{\boldsymbol{p}_h} \|_1 \Big)
\end{equation}
where $\boldsymbol{o}_{\boldsymbol{p}} \in {(0, 1)}^{K}$ represents the object assignment of center point $\boldsymbol{p}$, and $\boldsymbol{o}_{\boldsymbol{p}_h} \in {(0, 1)}^{K}$ represents the object assignment of its $h^{th}$ neighbouring point. This loss aims to avoid the over-segmentation issues. More details are provided in \cite{Song2022}.

In our experiments for DefGS and DefGS$_{nvfi}$, the maximum number of predicted objects $K$ is set to be 8. A softmax activation is applied on per-Gaussian object codes. During optimization, we adopt the Adam optimizer with a learning rate of 0.01 and optimize object codes for 1000 iterations until convergence.

All quantitative results for scene decomposition are in Table \ref{tab:segmentaion_all}.

\begin{table*}\tabcolsep=0.2cm 
\centering
\caption{Quantitative results of scene decomposition on the Synthetic Indoor Scene dataset.}
\label{tab:segmentaion_all}
\resizebox{1.0\linewidth}{!}{
\begin{tabular}{r|cccccc|cccccc}
\hline
& \multicolumn{6}{c|}{Gnome House} & \multicolumn{6}{c}{Chessboard} \\ \cline{2-13}
            & AP$\uparrow$    & PQ$\uparrow$    & F1$\uparrow$    & Pre$\uparrow$    & Rec$\uparrow$    & mIoU$\uparrow$ & AP$\uparrow$    & PQ$\uparrow$    & F1$\uparrow$    & Pre$\uparrow$    & Rec$\uparrow$    & mIoU$\uparrow$  \\ \hline
Mask2Former \cite{Cheng2022} & 60.89 & 73.05 & 77.32 & 85.32 & 70.69 & 66.94 & \underline{82.68} & \underline{81.35} & \underline{90.81} & \underline{97.54} & \underline{84.94} & \underline{76.17}\\
D-NeRF \cite{Pumarola2021} & 80.54 & 62.24 & 85.28 & 85.28 & 85.28 & 54.82 & 57.12 & 48.11 & 60.22 & 56.20 & 64.85 & 48.97\\
NVFi \citep{Li2023c} & \textbf{100.00} & 85.01 & \textbf{100.00} & \textbf{100.00} & \textbf{100.00} & 68.01 & 67.97 & 57.95 & 76.96 & 76.96 & 76.96 & 56.79\\
DefGS \citep{Yang2024} & 86.67 & 86.04 & 91.44 & 85.91 & 97.74 & 74.21 & 42.90 & 49.48 & 60.75 & 56.53 & 65.64 & 50.29\\
DefGS$_{nvfi}$ & \underline{99.12} & \underline{96.02} & \underline{99.17} & \underline{98.36} & \textbf{100.00} & \underline{77.45} & 31.27 & 47.55 & 54.87 & 56.87 & 53.01 & 44.41\\ \hline
\textbf{\nickname{} (Ours)} & \textbf{100.00} & \textbf{97.59} & \textbf{100.00} & \textbf{100.00} & \textbf{100.00} & \textbf{78.07} & \textbf{100.00} & \textbf{92.83} & \textbf{100.00} & \textbf{100.00} & \textbf{100.00} & \textbf{79.57}\\ \hline
& \multicolumn{6}{c|}{Dining Table} & \multicolumn{6}{c}{Factory} \\ \cline{2-13}
            & AP$\uparrow$    & PQ$\uparrow$    & F1$\uparrow$    & Pre$\uparrow$    & Rec$\uparrow$    & mIoU$\uparrow$ & AP$\uparrow$    & PQ$\uparrow$    & F1$\uparrow$    & Pre$\uparrow$    & Rec$\uparrow$    & mIoU$\uparrow$  \\ \hline
Mask2Former \cite{Cheng2022} & 77.65 & 84.61 & 87.42 & 97.44 & 79.28 & \underline{76.80} & 40.25 & 53.54 & 57.60 & 99.01 & 40.61 & 37.76\\
D-NeRF \cite{Pumarola2021} & 74.05 & 57.15 & 69.3 & 59.35 & 83.27 & 61.82 & 17.33 & 17.08 & 21.29 & 25.35 & 18.35 & 20.72\\
NVFi \citep{Li2023c} & \underline{98.01} & \underline{91.81} & \underline{98.95} & \underline{98.99} & 98.92 & 76.68 & \underline{98.86} & \underline{80.17} & \underline{99.09} & \underline{99.09} & \underline{99.09} & \underline{69.07}\\
DefGS \citep{Yang2024} & 57.66 & 62.92 & 70.51 & 69.12 & 71.95 & 55.73 & 19.69 & 31.96 & 43.02 & 41.27 & 44.94 & 37.59\\
DefGS$_{nvfi}$ & 67.02 & 72.12 & 78.37 & 64.85 & \underline{99.01} & 76.19 & 23.64 & 35.30 & 46.90 & 57.49 & 39.60 & 29.23\\ \hline
\textbf{\nickname{} (Ours)} & \textbf{98.99} & \textbf{98.36} & \textbf{99.97} & \textbf{99.98} & \textbf{99.97} & \textbf{81.89} & \textbf{100.00} & \textbf{96.31} & \textbf{100.00} & \textbf{100.00} & \textbf{100.00} & \textbf{82.55}\\ \hline
\end{tabular}
}
% \vspace{-0.4cm}
\end{table*}

\section{More Results of Ablation Study}
\label{app:ablation_all}
We report all ablation results in Table \ref{tab:ablation_all}. We conduct all ablations at a setting of $K=16$ originally, while we find $K=32$ is slightly better on Dynamic Indoor Scene dataset.
Nevertheless, this does not influence the analysis to the influencing factors as shown in the main paper. 

\begin{table*}[t]\tabcolsep=0.4cm 
\centering
\caption{Complete ablation study results on both Dynamic Object Dataset and Dynamic Indoor Scene Dataset}
\resizebox{1\linewidth}{!}{
\begin{tabular}{ccccc|ccc|ccc}
\hline
\multirow{3}{*}{} & \multirow{3}{*}{} & \multirow{3}{*}{} & \multirow{3}{*}{} & \multirow{3}{*}{} & \multicolumn{6}{c}{Dynamic Object Dataset} \\ \cline{6-11} 
& & & & & \multicolumn{3}{c|}{Interpolation} & \multicolumn{3}{c}{Extrapolation} \\ \cline{1-11} 
  & Code $\z$ & $f_{deform}$ & $\vel(\p_t,t)$ & $K$ & PSNR$\uparrow$ & SSIM$\uparrow$ & \multicolumn{1}{c|}{LPIPS$\downarrow$} & PSNR$\uparrow$ & SSIM$\uparrow$ & LPIPS$\downarrow$  \\ 
 \hline 
(1) & learnable & full & full & 16 & 38.722 & \underline{0.995} &  \underline{0.005}  & 25.961 & 0.975 & 0.025 \\ \hline
(2) & field & full & w/o $\mathcal{B}(\p_t)$ & 16  & 39.126 & \underline{0.995} & \underline{0.005} & 29.400 & 0.986 & 0.010 \\
(3) & field & full & w/o decomp & 16  & 39.111 & \underline{0.995} &  \underline{0.005} & 29.432 & 0.985 & \underline{0.009} \\
(4) & field & full & full & 8  & \underline{39.324 }& \textbf{0.996} & \textbf{0.004} & 30.972 & \underline{0.989} & \underline{0.009} \\
(4) & field & full & full & 32  & 39.318 & \textbf{0.996} & \textbf{0.004} & 31.438 & \textbf{0.990} & \textbf{0.007} \\ \hline
(5) & field & \N & full & 16  & 20.974 & 0.945 & 0.068 & 17.927 & 0.922 & 0.088 \\ 
(6) & field & w/o $\z$ & full & 16  & 39.151 & \underline{0.995} & \underline{0.005} & 31.217 & 0.988 & \underline{0.009} \\ 
(7) & field & w/o $\delta\s$ & full & 16  & 39.191 & \underline{0.995} & \underline{0.005} & \underline{31.704} & \textbf{0.990} & \textbf{0.007} \\ \hline
\textbf{\nickname{}} & field & full & full & 16  & \textbf{39.393} & \underline{0.995} & \underline{0.005} & \textbf{31.987} & \textbf{0.990} & \textbf{0.007} \\ [+0.1em] 
\hline
\multirow{3}{*}{} & \multirow{3}{*}{} & \multirow{3}{*}{} & \multirow{3}{*}{} & \multirow{3}{*}{} & \multicolumn{6}{c}{Dynamic Indoor Scene Dataset} \\ \cline{6-11} 
& & & & & \multicolumn{3}{c|}{Interpolation} & \multicolumn{3}{c}{Extrapolation} \\ \cline{1-11} 
  & Code $\z$ & $f_{deform}$ & $\vel(\p_t,t)$ & $K$ & PSNR$\uparrow$ & SSIM$\uparrow$ & \multicolumn{1}{c|}{LPIPS$\downarrow$} & PSNR$\uparrow$ & SSIM$\uparrow$ & LPIPS$\downarrow$  \\ 
 \hline 
(1) & learnable & full & full & 16 & \textbf{32.343} & \textbf{0.930} &  \textbf{0.091} & 30.444 & 0.933 & 0.087 \\ \hline
(2) & field & full & w/o $\mathcal{B}(\p_t)$ & 16  & 31.471 & 0.921 & 0.108& 32.316 & 0.944 & 0.077 \\
(3) & field & full & w/o decomp & 16  & 31.707 & 0.921 & 0.106 & 31.204 & 0.943 & 0.071 \\
(4) & field & full & full & 8 & 32.005 & \underline{0.929} & 0.093 & 34.159 & 0.962 & 0.053 \\
(4) & field & full & full & 16 & 31.996 & \underline{0.929 }& \underline{0.092} & \underline{34.716} & \underline{0.965} & \textbf{0.051} \\ \hline
(5) & field & \N & full & 16 & - & - & - & - & - & - \\ 
(6) & field & w/o $\z$ & full & 16 & 31.603 & 0.921 & 0.107 & 33.408 & 0.955 & 0.067 \\ 
(7) & field & w/o $\delta\s$ & full & 16 & 32.094 & \underline{0.929} & \underline{0.092} & 34.504 & 0.964 & \underline{0.052} \\ \hline
\textbf{\nickname{}} & field & full & full & 32  & \underline{32.287} & \textbf{0.930} & \underline{0.092} & \textbf{35.019} & \textbf{0.966} & \textbf{0.051} \\ [+0.1em] 
\hline\\ [+0.1em] 
\end{tabular}
}
\label{tab:ablation_all}
%\vspace{-0.4cm}
\end{table*}

\section{More Results on Dynamic Object Dataset}
\label{app:results_object}
The quantitative results for each scene of Dynamic Object Dataset are in Table \ref{tab:object_all}. 

\begin{table*}[t]\tabcolsep=0.2cm  
\centering
\caption{Per-scene quantitative results on Dynamic Object Dataset.}
\resizebox{1.0\linewidth}{!}{
\begin{tabular}{r|ccc|ccc|ccc|ccc}
\hline
 & \multicolumn{6}{c|}{Falling Ball} & \multicolumn{6}{c}{Bat} \\ \cline{2-13} 
Methods & \multicolumn{3}{c|}{Interpolation} & \multicolumn{3}{c|}{Extrapolation} & \multicolumn{3}{c|}{Interpolation} & \multicolumn{3}{c}{Extrapolation} \\ \cline{2-13} 
 & PSNR$\uparrow$ & SSIM$\uparrow$ & LPIPS$\downarrow$ & PSNR$\uparrow$ & SSIM$\uparrow$ & LPIPS$\downarrow$ & PSNR$\uparrow$ & SSIM$\uparrow$ & LPIPS$\downarrow$ & PSNR$\uparrow$ & SSIM$\uparrow$ & LPIPS$\downarrow$ \\ \hline
T-NeRF \citep{Pumarola2021} & 14.921 & 0.782 & 0.326 & 15.418 & 0.793 & 0.308 & 13.070 & 0.836 & 0.234 & 13.897 & 0.834 & 0.230 \\
D-NeRF \citep{Pumarola2021} & 15.548 & 0.665 & 0.435 & 15.116 & 0.644 & 0.427 & 14.087 & 0.845 & 0.212 & 15.406 & 0.887 & 0.175 \\
TiNeuVox \citep{Fang2022} & 35.458 & 0.974 & 0.052 & 20.242 & 0.959 & 0.067 & 16.080 & 0.908 & 0.108 & 16.952 & 0.930 & 0.115 \\
T-NeRF$_{PINN}$ & 17.687 & 0.775 & 0.368 & 17.857 & 0.829 & 0.265 & 16.412 & 0.903 & 0.197 & 18.983 & 0.930 & 0.132 \\
HexPlane$_{PINN}$ & 32.144 & 0.965 & 0.065 & 20.762 & 0.951 & 0.081 & 23.399 & 0.958 & 0.057 & 21.144 & 0.951 & 0.064 \\ 
NVFi \citep{Li2023c}  & 35.826 & 0.978 & 0.041 & 31.369 & 0.978 & 0.041 & 23.325 & \underline{0.964} & 0.046 & 25.015 & 0.968 & 0.042 \\ 
DefGS \citep{Yang2024} & 37.535 & 0.995 & \underline{0.009} & 20.442 & 0.976 & 0.033 & \underline{38.750} & \textbf{0.997} & \underline{0.004} & 17.063 & 0.936 & 0.072 \\ 
DefGS$_{nvfi}$ & \underline{38.606} & \underline{0.996} & 0.010 & \underline{24.873} & \underline{0.985} & \underline{0.015} & 38.075 & \textbf{0.997} & \underline{0.004} & \textbf{28.950} & \underline{0.980} & \textbf{0.015} \\ 
\hline
\textbf{\nickname{} (Ours)} & \textbf{42.369} & \textbf{0.998} & \textbf{0.003}& \textbf{38.321} & \textbf{0.997} & \textbf{0.003} & \textbf{39.662} & \textbf{0.997} & \textbf{0.002} & \underline{27.235} & \textbf{0.982} & \underline{0.013} \\ \hline

 & \multicolumn{6}{c|}{Fan} & \multicolumn{6}{c}{Telescope} \\ \cline{2-13} 
Methods & \multicolumn{3}{c|}{Interpolation} & \multicolumn{3}{c|}{Extrapolation} & \multicolumn{3}{c|}{Interpolation} & \multicolumn{3}{c}{Extrapolation} \\ \cline{2-13} 
 & PSNR$\uparrow$ & SSIM$\uparrow$ & LPIPS$\downarrow$ & PSNR$\uparrow$ & SSIM$\uparrow$ & LPIPS$\downarrow$ & PSNR$\uparrow$ & SSIM$\uparrow$ & LPIPS$\downarrow$ & PSNR$\uparrow$ & SSIM$\uparrow$ & LPIPS$\downarrow$ \\ \hline
T-NeRF \cite{Pumarola2021} & 8.001 & 0.308 & 0.646 & 8.494 & 0.392 & 0.593 & 13.031 & 0.615 & 0.472 & 13.892 & 0.670 & 0.417 \\
D-NeRF \cite{Pumarola2021} & 7.915 & 0.262 & 0.690 & 8.624 & 0.370 & 0.623 & 13.295 & 0.609 & 0.469 & 14.967 & 0.700 & 0.385 \\
TiNeuVox \cite{Fang2022} & 24.088 & 0.930 & 0.104 & 20.932 & 0.935 & 0.078 & 31.666 & 0.982 & 0.041 & 20.456 & 0.921 & 0.067 \\
T-NeRF$_{PINN}$ & 9.233 & 0.541 & 0.508 & 9.828 & 0.606 & 0.443 & 14.293 & 0.739 & 0.366 & 15.752 & 0.804 & 0.298 \\
HexPlane$_{PINN}$ & 22.822 & 0.921 & 0.079 & 19.724 & 0.919 & 0.080 & 25.381 & 0.948 & 0.066 & 23.165 & 0.932 & 0.074 \\ 
NVFi \cite{Li2023c}  & 25.213 & 0.948 & 0.049 & \underline{27.172} & 0.963 & 0.037 & 26.487 & 0.959 & 0.048 & 27.101 & 0.963 & 0.046 \\ 
DefGS \cite{Yang2024} & \textbf{35.858} & \textbf{0.985} & \underline{0.017} & 20.932 & 0.948 & 0.038 & 37.502 & \underline{0.996} & \underline{0.003} & 20.684 & 0.927 & 0.048 \\ 
DefGS$_{nvfi}$ & 35.217 & \underline{0.984} & 0.019 & 26.648 & \underline{0.972} & \underline{0.023} & \underline{37.568} & \underline{0.996} & \underline{0.003} & \underline{34.096} & \underline{0.994} & \underline{0.005} \\ 
\hline
\textbf{\nickname{} (Ours)} &  \underline{35.767} & \textbf{0.985} & \textbf{0.013} & \textbf{32.393} & \textbf{0.986} & \textbf{0.009} & \textbf{40.332} & \textbf{0.998} & \textbf{0.002} & \textbf{40.401} & \textbf{0.998} & \textbf{0.002}\\ 
\hline

 & \multicolumn{6}{c|}{Shark} & \multicolumn{6}{c}{Whale} \\ \cline{2-13} 
Methods & \multicolumn{3}{c|}{Interpolation} & \multicolumn{3}{c|}{Extrapolation} & \multicolumn{3}{c|}{Interpolation} & \multicolumn{3}{c}{Extrapolation} \\ \cline{2-13} 
 & PSNR$\uparrow$ & SSIM$\uparrow$ & LPIPS$\downarrow$ & PSNR$\uparrow$ & SSIM$\uparrow$ & LPIPS$\downarrow$ & PSNR$\uparrow$ & SSIM$\uparrow$ & LPIPS$\downarrow$ & PSNR$\uparrow$ & SSIM$\uparrow$ & LPIPS$\downarrow$ \\ \hline
T-NeRF \cite{Pumarola2021} & 13.813 & 0.853 & 0.223 & 15.325 & 0.882 & 0.193 & 16.141 & 0.860 & 0.212 & 15.880 & 0.860 & 0.203 \\
D-NeRF \cite{Pumarola2021} & 17.727 & 0.903 & 0.150 & 19.078 & 0.936 & 0.092 & 16.373 & 0.898 & 0.154 & 14.771 & 0.883 & 0.171 \\
TiNeuVox \cite{Fang2022} & 23.178 & 0.971 & 0.059 & 19.463 & 0.950 & 0.050 & 37.455 & 0.994 & 0.016 & 19.624 & 0.943 & 0.063 \\
T-NeRF$_{PINN}$ & 17.315 & 0.878 & 0.177 & 18.739 & 0.921 & 0.115 & 16.778 & 0.927 & 0.141 & 15.974 & 0.919 & 0.127 \\
HexPlane$_{PINN}$ & 28.874 & 0.976 & 0.040 & 22.330 & 0.961 & 0.047 & 29.634 & 0.981 & 0.035 & 21.391 & 0.961 & 0.053 \\ 
NVFi \cite{Li2023c}  & 32.072 & 0.984 & 0.024 & 28.874 & 0.982 & 0.021 & 31.240 & 0.986 & 0.025 & 26.032 & 0.978 & 0.029 \\ 
DefGS \cite{Yang2024} & \underline{37.802} & \underline{0.994} & \underline{0.006} & 19.924 & 0.957 & 0.034 & \textbf{39.740} & \textbf{0.997} & \underline{0.004} & 20.048 & 0.951 & 0.046 \\ 
DefGS$_{nvfi}$ & 37.327 & \underline{0.994} & \underline{0.006} & \textbf{29.240} & \underline{0.987} & \underline{0.007} & 37.101 & \underline{0.996} & 0.005 & \underline{28.686} & \underline{0.986} & \underline{0.012} \\ 
\hline
\textbf{\nickname{} (Ours)} & \textbf{40.211} & \textbf{0.996} & \textbf{0.004} & \underline{29.236} & \textbf{0.990} & \textbf{0.005} & \underline{38.015} & \textbf{0.997} & \textbf{0.003} & \textbf{28.950} & \textbf{0.989} & \textbf{0.009} \\ 
\hline
\end{tabular}
}
\label{tab:object_all}
\vspace{-0.1cm}
\end{table*}

\section{More Results on Dynamic Indoor Scene Dataset}

The quantitative results for each scene of Dynamic Indoor Scene Dataset are in Table \ref{tab:indoor_all}. 

\begin{table*}[t]\tabcolsep=0.2cm 
\footnotesize
\centering
\caption{Per-scene quantitative results on Dynamic Indoor Scene Dataset.}
\resizebox{1.0\linewidth}{!}{
\begin{tabular}{r|ccc|ccc|ccc|ccc}
\hline
 & \multicolumn{6}{c|}{Gnome House} & \multicolumn{6}{c}{Chessboard} \\ \cline{2-13} 
Methods & \multicolumn{3}{c|}{Interpolation} & \multicolumn{3}{c|}{Extrapolation} & \multicolumn{3}{c|}{Interpolation} & \multicolumn{3}{c}{Extrapolation} \\ \cline{2-13} 
 & PSNR$\uparrow$ & SSIM$\uparrow$ & LPIPS$\downarrow$ & PSNR$\uparrow$ & SSIM$\uparrow$ & LPIPS$\downarrow$ & PSNR$\uparrow$ & SSIM$\uparrow$ & LPIPS$\downarrow$ & PSNR$\uparrow$ & SSIM$\uparrow$ & LPIPS$\downarrow$ \\ \hline
T-NeRF \citep{Pumarola2021} & 26.094 & 0.716 & 0.383 & 23.485 & 0.643 & 0.419 & 25.517 & 0.796 & 0.294 & 20.228 & 0.708 & 0.365 \\
D-NeRF \citep{Pumarola2021} & 27.000 & 0.745 & 0.319 & 21.714 & 0.641 & 0.367 & 24.852 & 0.774 & 0.308 & 19.455 & 0.675 & 0.384 \\
TiNeuVox \citep{Fang2022} & 30.646 & 0.831 & 0.253 & 21.418 & 0.699 & 0.326 & \underline{33.001} & \underline{0.917} & 0.177 & 19.718 & 0.765 & 0.310 \\
T-NeRF$_{PINN}$ & 15.008 & 0.375 & 0.668 & 16.200 & 0.409 & 0.651 & 16.549 & 0.457 & 0.621 & 17.197 & 0.472 & 0.618 \\
HexPlane$_{PINN}$ & 23.764 & 0.658 & 0.510 & 22.867 & 0.658 & 0.510 & 24.605 & 0.778 & 0.412 & 21.518 & 0.748 & 0.428 \\ 
NSFF \citep{Li2021c} & 31.418 & 0.821 & 0.294 & 25.892 & 0.750 & 0.327 & 32.514 & 0.810 & 0.201 & 21.501 & 0.805 & 0.282 \\ 
NVFi \citep{Li2023c}  & 30.667 & 0.824 & 0.277 & 30.408 & 0.826 & 0.273 & 30.394 & 0.888 & 0.215 & 27.840 & 0.872 & 0.219 \\
DefGS \citep{Yang2024} & 32.041 & 0.918 & \underline{0.132} & 21.703 & 0.775 & 0.207 & 27.355 & 0.912 & \underline{0.147} & 20.032 & 0.808 & 0.218 \\ 
DefGS$_{nvfi}$ & \textbf{32.881} & \underline{0.919} & \underline{0.132} & \underline{33.630} & \underline{0.953} & \underline{0.077} & 26.200 & 0.907 & 0.156 & \underline{26.730} & \underline{0.917} & \underline{0.110} \\ 
\hline
\textbf{\nickname{} (Ours)} & \underline{32.791} & \textbf{0.923} & \textbf{0.103} & \textbf{36.458}  & \textbf{0.963} & \textbf{0.062} & \textbf{35.388} & \textbf{0.962} & \textbf{0.061} & \textbf{35.016} & \textbf{0.970} & \textbf{0.044} \\ 
\hline

 & \multicolumn{6}{c|}{Factory} & \multicolumn{6}{c}{Dining Table} \\ \cline{2-13} 
Methods & \multicolumn{3}{c|}{Interpolation} & \multicolumn{3}{c|}{Extrapolation} & \multicolumn{3}{c|}{Interpolation} & \multicolumn{3}{c}{Extrapolation} \\ \cline{2-13} 
 & PSNR$\uparrow$ & SSIM$\uparrow$ & LPIPS$\downarrow$ & PSNR$\uparrow$ & SSIM$\uparrow$ & LPIPS$\downarrow$ & PSNR$\uparrow$ & SSIM$\uparrow$ & LPIPS$\downarrow$ & PSNR$\uparrow$ & SSIM$\uparrow$ & LPIPS$\downarrow$ \\ \hline
T-NeRF \cite{Pumarola2021} & 26.467 & 0.741 & 0.328 & 24.276 & 0.722 & 0.344 & 21.699 & 0.716 & 0.338 & 20.977 & 0.725 & 0.324 \\
D-NeRF \cite{Pumarola2021} & 28.818 & 0.818 & 0.252 & 22.959 & 0.746 & 0.303 & 20.851 & 0.725 & 0.319 & 19.035 & 0.705 & 0.341 \\
TiNeuVox \cite{Fang2022} & 32.684 & 0.909 & 0.148 & 22.622 & 0.810 & 0.229 & 23.596 & 0.798 & 0.274 & 20.357 & 0.804 & 0.258 \\
T-NeRF$_{PINN}$ & 16.634 & 0.446 & 0.624 & 17.546 & 0.480 & 0.609 & 16.807 & 0.486 & 0.640 & 18.215 & 0.548 & 0.595 \\
HexPlane$_{PINN}$ & 27.200 & 0.826 & 0.283 & 24.998 & 0.792 & 0.312 & 25.291 & 0.788 & 0.350 & 22.979 & 0.771 & 0.355 \\ 
NSFF \cite{Li2021c} & \textbf{33.975} & \underline{0.919} & 0.152 & 26.647 & 0.855 & 0.196 & 19.552 & 0.665 & 0.464 & 22.612 & 0.770 & 0.351 \\ 
NVFi \cite{Li2023c}  & 32.460 & 0.912 & 0.151 & 31.719 & 0.908 & 0.154 & \textbf{29.179} & 0.885 & 0.199 & 29.011 & 0.898 & 0.171 \\ 
DefGS \cite{Yang2024} & 33.629 & \textbf{0.943} & \underline{0.096} & 22.820 & 0.839 & 0.169 & 27.680 & 0.890 & \underline{0.145} & 20.965 & 0.855 & 0.157 \\ 
DefGS$_{nvfi}$ & \underline{33.643} & \textbf{0.943} & 0.097 & \underline{33.049} & \underline{0.954} & \underline{0.062} & \underline{27.957} & \underline{0.891} & \underline{0.145} & \underline{30.975} & \underline{0.955} & \underline{0.060} \\ 
\hline
\textbf{\nickname{} (Ours)} & 33.316 & \textbf{0.943} & \textbf{0.079} & \textbf{35.765} & \textbf{0.966} & \textbf{0.048} & 27.652 & \textbf{0.892} & \textbf{0.124} & \textbf{32.838} & \textbf{0.963} & \textbf{0.048}  \\ \hline
\end{tabular}
}
\label{tab:indoor_all}
\vspace{-0.1cm}
\end{table*}

\section{More Results on \nickname-GoPro Dataset}
The quantitative results for each scene of \nickname-GoPro Dataset are in Table \ref{tab:gopro_all}. 

\begin{table*}[t]\tabcolsep=0.2cm 
\footnotesize
\centering
\caption{Per-scene quantitative results on \nickname-GoPro Dataset.}
\resizebox{1.0\linewidth}{!}{
\begin{tabular}{r|ccc|ccc|ccc|ccc}
\hline
 & \multicolumn{6}{c|}{Pen \& Tape 1} & \multicolumn{6}{c}{Pen \& Tape 2} \\ \cline{2-13} 
Methods & \multicolumn{3}{c|}{Interpolation} & \multicolumn{3}{c|}{Extrapolation} & \multicolumn{3}{c|}{Interpolation} & \multicolumn{3}{c}{Extrapolation} \\ \cline{2-13} 
 & PSNR$\uparrow$ & SSIM$\uparrow$ & LPIPS$\downarrow$ & PSNR$\uparrow$ & SSIM$\uparrow$ & LPIPS$\downarrow$ & PSNR$\uparrow$ & SSIM$\uparrow$ & LPIPS$\downarrow$ & PSNR$\uparrow$ & SSIM$\uparrow$ & LPIPS$\downarrow$ \\ \hline
TiNeuVox \citep{Fang2022} & 19.368 & 0.758 & 0.304 & 20.127 & 0.795 & 0.289 & 19.594 & 0.732 & 0.334 & 20.514 & 0.779 & 0.296 \\
NVFi \citep{Li2023c}  & 21.397 & 0.816 & 0.243 & 23.869 & 0.824 & 0.258 & 22.31 & 0.813 & 0.245 & 23.574 & 0.806 & 0.269   \\ 
DefGS \citep{Yang2024} & \textbf{29.598} & \textbf{0.933} & \textbf{0.080} & 20.284 & 0.865 & 0.163 & \textbf{27.587} & \textbf{0.909} & \textbf{0.098} & 20.674 & 0.861 & 0.169  \\ 
DefGS$_{nvfi}$ & \underline{29.571} & \underline{0.932} & \underline{0.081} & \underline{26.289} & \underline{0.922} & \underline{0.108} & 27.456 & \textbf{0.909} & \underline{0.099} & \underline{27.124} & \underline{0.915} & \underline{0.120}  \\ 
\hline
\textbf{\nickname{} (Ours)} & 29.412 & 0.927 & 0.087 & \textbf{29.001} & \textbf{0.936} & \textbf{0.090} & \underline{27.498} & \underline{0.907} & \textbf{0.098} & \textbf{28.842} & \textbf{0.925} & \textbf{0.107}  \\ \hline

 & \multicolumn{6}{c|}{Box} & \multicolumn{6}{c}{Wrist Rest} \\ \cline{2-13} 
Methods & \multicolumn{3}{c|}{Interpolation} & \multicolumn{3}{c|}{Extrapolation} & \multicolumn{3}{c|}{Interpolation} & \multicolumn{3}{c}{Extrapolation} \\ \cline{2-13} 
 & PSNR$\uparrow$ & SSIM$\uparrow$ & LPIPS$\downarrow$ & PSNR$\uparrow$ & SSIM$\uparrow$ & LPIPS$\downarrow$ & PSNR$\uparrow$ & SSIM$\uparrow$ & LPIPS$\downarrow$ & PSNR$\uparrow$ & SSIM$\uparrow$ & LPIPS$\downarrow$ \\ \hline
TiNeuVox \cite{Fang2022} & 19.464 & 0.726 & 0.318 & 23.958 & 0.807 & 0.247 & 19.133 & 0.751 & 0.315 & 18.204 & 0.727 & 0.342   \\
NVFi \cite{Li2023c}  & 19.391 & 0.777 & 0.282 & 24.867 & 0.806 & 0.263 & 13.235 & 0.570 & 0.490 & 19.222 & 0.683  & 0.431  \\ 
DefGS \cite{Yang2024} & \underline{28.448} & \underline{0.918} & \underline{0.087} & 25.656 & 0.904 & 0.117 & \underline{28.178} & \underline{0.923} & \underline{0.100} & 18.834 & 0.809 & 0.220   \\ 
DefGS$_{nvfi}$ & \textbf{29.571} & \textbf{0.932 }& \textbf{0.081} & \underline{26.289 }& \underline{0.922} & \underline{0.108} & 27.938 & 0.921 & 0.102 & \underline{22.741} & \underline{0.856} & \underline{0.170 }  \\ 
\hline
\textbf{\nickname{} (Ours)} & 28.339 & 0.916 & 0.088 & \textbf{30.964} & \textbf{0.935} & \textbf{0.084} & \textbf{28.708} & \textbf{0.925} & \textbf{0.098} & \textbf{24.093} & \textbf{0.867} & \textbf{0.159}   \\ 
\hline

 & \multicolumn{6}{c|}{Hammer} & \multicolumn{6}{c}{Collision} \\ \cline{2-13} 
Methods & \multicolumn{3}{c}{Interpolation} & \multicolumn{3}{c|}{Extrapolation} & \multicolumn{3}{c}{Interpolation} & \multicolumn{3}{c}{Extrapolation} \\ \cline{2-13} 
 & PSNR$\uparrow$ & SSIM$\uparrow$ &    LPIPS$\downarrow$ & PSNR$\uparrow$ & SSIM$\uparrow$ & LPIPS$\downarrow$ & PSNR$\uparrow$ & SSIM$\uparrow$ & LPIPS$\downarrow$ & PSNR$\uparrow$ & SSIM$\uparrow$ & LPIPS$\downarrow$ \\ \hline
TiNeuVox \cite{Fang2022} & 18.75 & 0.733 & 0.324 & 22.638 & 0.710 & 0.251  & 17.848 & 0.741 & 0.316 & 18.158 & 0.743 & 0.329 \\
NVFi \cite{Li2023c}  & 22.817 & 0.817 & 0.235 & 25.526 & 0.830 & 0.241 & 14.530 & 0.638 & 0.438 & 19.422 & 0.717 & 0.391  \\ 
DefGS \cite{Yang2024} & 28.141 & \underline{0.916} &\textbf{ 0.089 } & 23.995 & 0.899 & 0.123 & \textbf{28.493} & \underline{0.923} & \underline{ 0.089} & 18.619 & 0.808 & 0.228  \\ 
DefGS$_{nvfi}$ & \textbf{28.478} & \textbf{0.917 }& \underline{0.088} & \underline{29.392} & \underline{0.928} & \underline{0.095} & 28.125 & \underline{0.923} & 0.092 & \underline{ 23.512} & \underline{0.871} & \underline{0.157}   \\ 
\hline
\textbf{\nickname{} (Ours)} & \underline{28.314} & \textbf{0.917} & \textbf{0.089} & \textbf{30.090} & \textbf{0.932 }& \textbf{0.091} & \underline{28.434} & \textbf{0.925} & \textbf{0.088} & \textbf{25.571} & \textbf{0.886 }& \textbf{0.139 }  \\ 
\hline
\end{tabular}
}
\label{tab:gopro_all}
\vspace{-0.1cm}
\end{table*}

\section{More Results on ParticleNeRF Dataset}
\label{app:results_particlenerf}
The quantitative results for each scene of ParticleNeRF Dataset are in Table \ref{tab:particlenerf_all}. 

\begin{table*}[t]\tabcolsep=0.2cm  
\centering
\caption{Per-scene quantitative results on ParticleNeRF Dataset.}
\resizebox{1.0\linewidth}{!}{
\begin{tabular}{r|ccc|ccc|ccc|ccc}
\hline
 & \multicolumn{6}{c|}{Robot} & \multicolumn{6}{c}{Robot Task} \\ \cline{2-13} 
Methods & \multicolumn{3}{c|}{Interpolation} & \multicolumn{3}{c|}{Extrapolation} & \multicolumn{3}{c|}{Interpolation} & \multicolumn{3}{c}{Extrapolation} \\ \cline{2-13} 
 & PSNR$\uparrow$ & SSIM$\uparrow$ & LPIPS$\downarrow$ & PSNR$\uparrow$ & SSIM$\uparrow$ & LPIPS$\downarrow$ & PSNR$\uparrow$ & SSIM$\uparrow$ & LPIPS$\downarrow$ & PSNR$\uparrow$ & SSIM$\uparrow$ & LPIPS$\downarrow$ \\ \hline
TiNeuVox \citep{Fang2022} & 32.079 & 0.975 & 0.063 & 17.287 & 0.861 & 0.162 & 34.672 & 0.984 & 0.047 & 21.078 & 0.906 & 0.097 \\
NVFi \citep{Li2023c}  &  28.740 & 0.962 & 0.065 & 18.518 & 0.875 & 0.125 & 30.906 & 0.971 & 0.051 & 26.130 & 0.945 & 0.067 \\ 
DefGS \citep{Yang2024} & 34.713 & \textbf{0.989} & \textbf{0.015} & 15.793 & 0.872 & 0.129 & 37.218 & \textbf{0.994} & \textbf{0.006} & 19.193 & 0.911 & 0.079 \\ 
DefGS$_{nvfi}$ & \textbf{33.924} & 0.987 & 0.017 & 17.965 & 0.892 & 0.092 & \textbf{37.640} & \textbf{0.994} & \textbf{0.006} & \textbf{26.566} & \textbf{0.962} & \textbf{0.023}\\ 
\hline
\textbf{\nickname{} (Ours)} & 33.298 & 0.986 & 0.017 & \textbf{19.361} & \textbf{0.901} & \textbf{0.076} & 37.538 & \textbf{0.994} & \textbf{0.006} & 25.526 & 0.954 & 0.029\\ \hline

 & \multicolumn{6}{c|}{Cloth} & \multicolumn{6}{c}{Wheel} \\ \cline{2-13} 
Methods & \multicolumn{3}{c|}{Interpolation} & \multicolumn{3}{c|}{Extrapolation} & \multicolumn{3}{c|}{Interpolation} & \multicolumn{3}{c}{Extrapolation} \\ \cline{2-13} 
 & PSNR$\uparrow$ & SSIM$\uparrow$ & LPIPS$\downarrow$ & PSNR$\uparrow$ & SSIM$\uparrow$ & LPIPS$\downarrow$ & PSNR$\uparrow$ & SSIM$\uparrow$ & LPIPS$\downarrow$ & PSNR$\uparrow$ & SSIM$\uparrow$ & LPIPS$\downarrow$ \\ \hline
TiNeuVox \cite{Fang2022} & 32.406 & 0.981 & 0.052 & 18.476 & 0.885 & 0.117 & 28.544 & 0.946 & 0.058 & 22.599 & 0.880 & 0.079\\
NVFi \cite{Li2023c}  & 27.309 & 0.951 & 0.075 & 18.904 & 0.894 & 0.116 & 26.225 & 0.935 & 0.056 & 12.990 & 0.790 & 0.153 \\ 
DefGS \cite{Yang2024} & \textbf{34.072} & \textbf{0.991} & \textbf{0.010} & 16.687 & 0.880 & 0.115 & 30.290 & \textbf{0.971} & \textbf{0.028} & 25.840 & 0.945 & 0.034\\ 
DefGS$_{nvfi}$ & 32.547 & 0.986 & 0.012 & 26.655 & 0.964 & \textbf{0.023} & 28.537 & 0.968 & 0.029 & 22.393 & 0.914 & 0.063\\ 
\hline
\textbf{\nickname{} (Ours)} & 32.604 & 0.987 & 0.011 & \textbf{27.934} & \textbf{0.966} & 0.026 & \textbf{30.350} & \textbf{0.971} & \textbf{0.028} & \textbf{30.926} & \textbf{0.972} & \textbf{0.022}\\ 
\hline

 & \multicolumn{6}{c|}{Spring} & \multicolumn{6}{c}{Pendulums} \\ \cline{2-13} 
Methods & \multicolumn{3}{c|}{Interpolation} & \multicolumn{3}{c|}{Extrapolation} & \multicolumn{3}{c|}{Interpolation} & \multicolumn{3}{c}{Extrapolation} \\ \cline{2-13} 
 & PSNR$\uparrow$ & SSIM$\uparrow$ & LPIPS$\downarrow$ & PSNR$\uparrow$ & SSIM$\uparrow$ & LPIPS$\downarrow$ & PSNR$\uparrow$ & SSIM$\uparrow$ & LPIPS$\downarrow$ & PSNR$\uparrow$ & SSIM$\uparrow$ & LPIPS$\downarrow$ \\ \hline
TiNeuVox \cite{Fang2022} & 32.731 & 0.990 & 0.022 & 20.448 & 0.891 & 0.073 & 36.093 & 0.991 & 0.028 & 22.551 & 0.905 & 0.084\\
NVFi \cite{Li2023c}  & 30.315 & 0.982 & 0.020 & 15.575 & 0.853 & 0.107 & 29.691 & 0.970 & 0.046 & 16.922 & 0.844 & 0.146\\ 
DefGS \cite{Yang2024} & 35.684 & 0.995 & 0.004 & 19.286 & 0.905 & 0.060 & \textbf{39.392} & \textbf{0.997} & \textbf{0.003} & 18.428 & 0.889 & 0.082\\ 
DefGS$_{nvfi}$ & 34.606 & 0.995 & 0.004 & 23.648 & 0.953 & 0.024 & 37.973 & 0.996 & 0.004 & 19.154 & 0.903 & 0.075\\ 
\hline
\textbf{\nickname{} (Ours)} & \textbf{38.465} & \textbf{0.997} & \textbf{0.003} & \textbf{25.501} & \textbf{0.959} & \textbf{0.015} & 38.992 & \textbf{0.997} & \textbf{0.003} & \textbf{30.696} & \textbf{0.985} & \textbf{0.009}\\ 
\hline
\end{tabular}
}
\label{tab:particlenerf_all}
% \vspace{-0.1cm}
\end{table*}

\section{Additional Qualitative Results}
\label{app:qualitative_all}

\begin{figure*}[t]
\setlength{\abovecaptionskip}{ 4. pt}
\setlength{\belowcaptionskip}{ -10 pt}
\centering
   \includegraphics[width=0.8\linewidth]{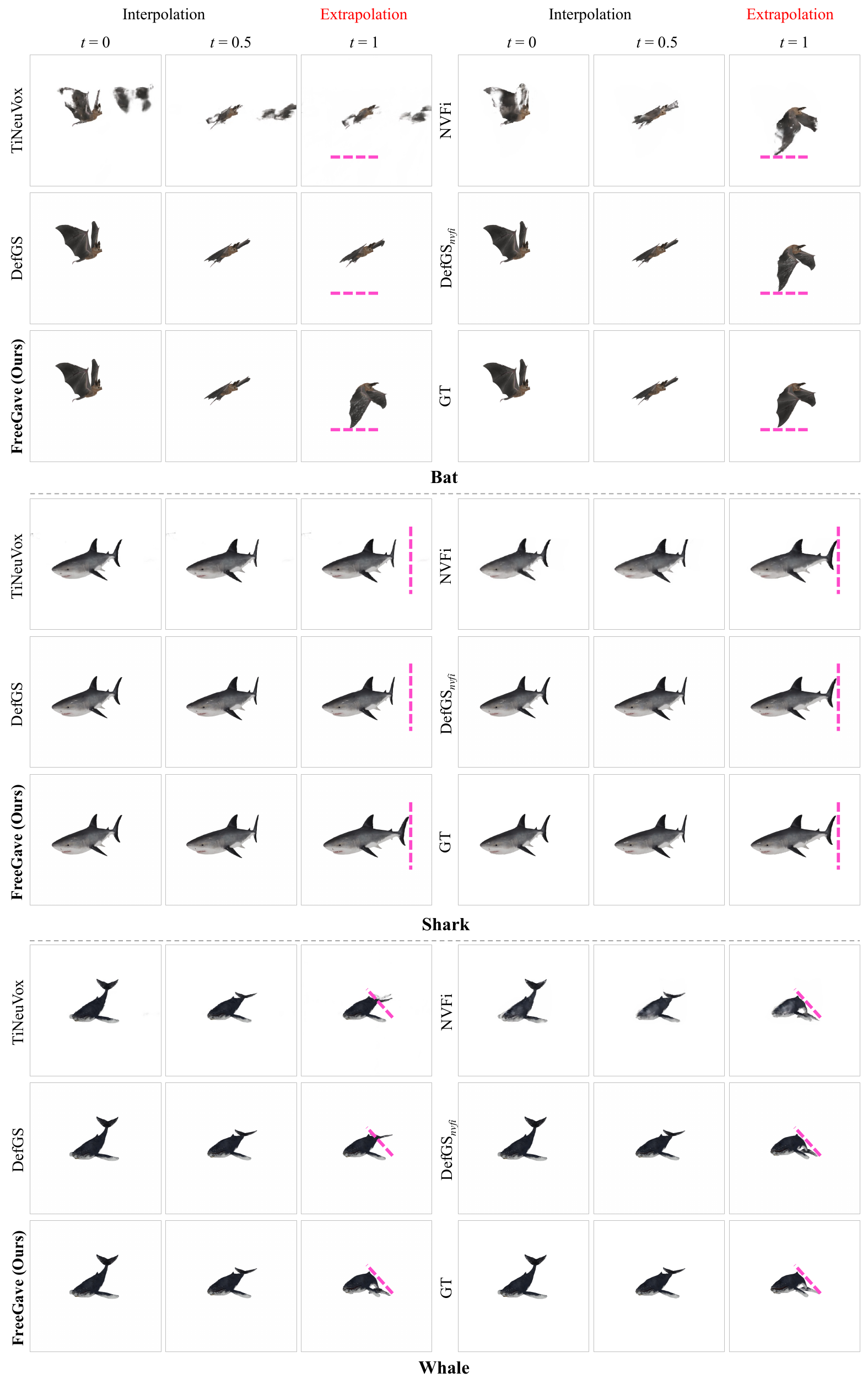}
\caption{Qualitative results for future frame extrapolation on Dynamic Object Dataset.}
\label{fig:qual_res_app1}
\end{figure*}

\begin{figure*}[t]
\setlength{\abovecaptionskip}{ 4. pt}
\setlength{\belowcaptionskip}{ -10 pt}
\centering
   \includegraphics[width=0.8\linewidth]{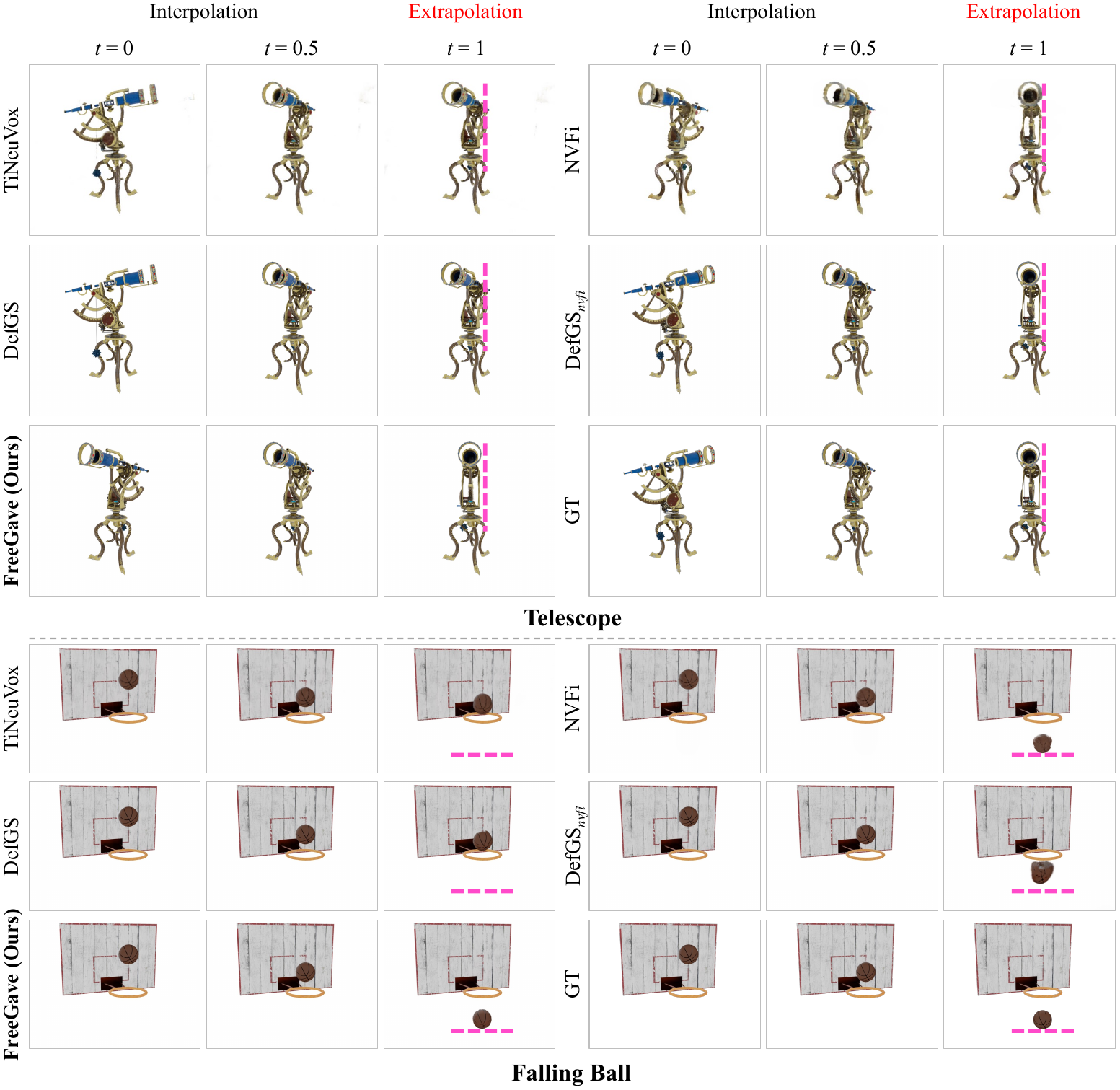}
\caption{Qualitative results for future frame extrapolation on Dynamic Object Dataset.}
\label{fig:qual_res_app2}
\end{figure*}

\begin{figure*}[t]
\setlength{\abovecaptionskip}{ 4. pt}
\setlength{\belowcaptionskip}{ -10 pt}
\centering
   \includegraphics[width=0.8\linewidth]{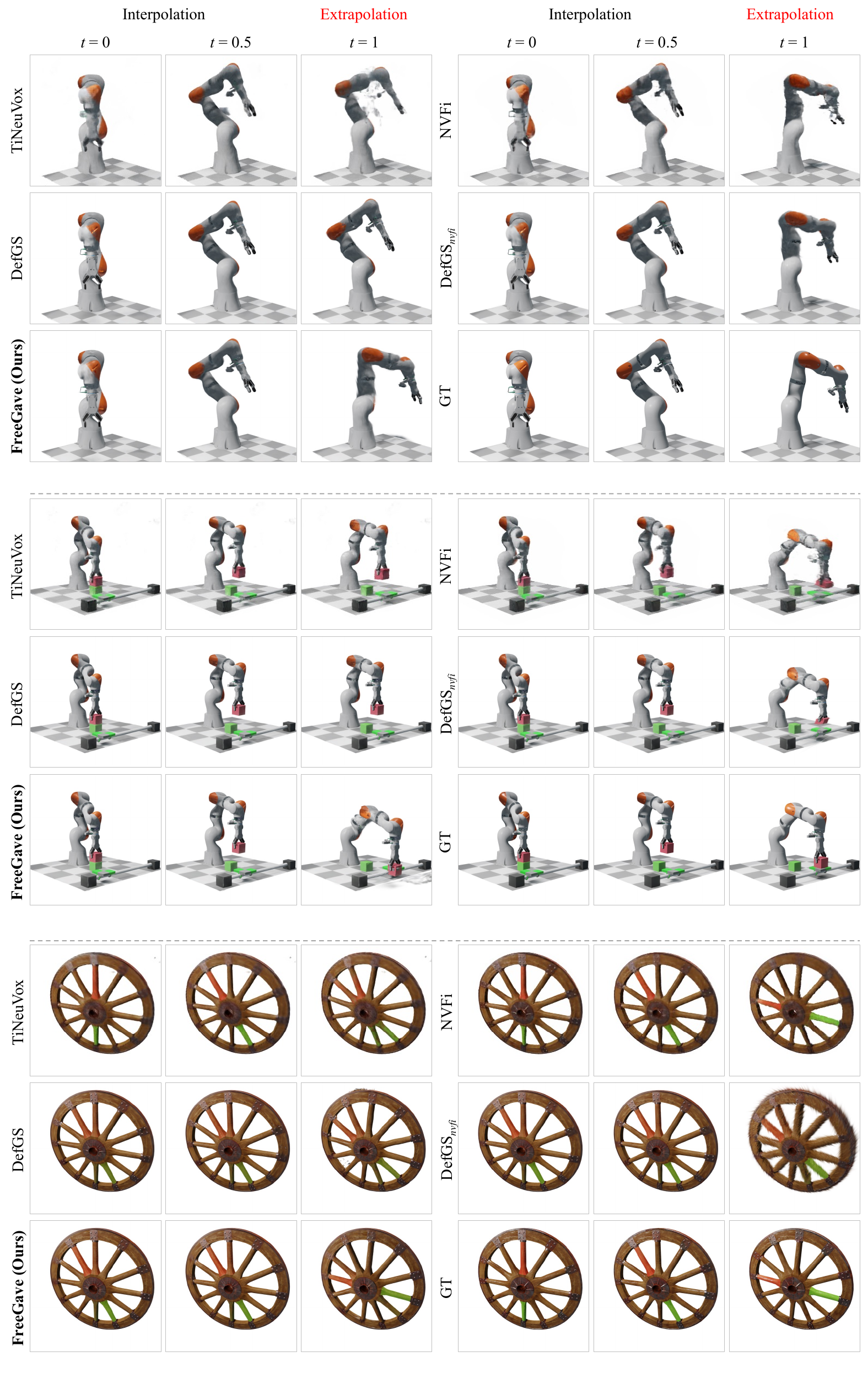}
\caption{Qualitative results for future frame extrapolation on ParticleNeRF Dataset.}
\label{fig:qual_res_app_particle1}
\end{figure*}

\begin{figure*}[t]
\setlength{\abovecaptionskip}{ 4. pt}
\setlength{\belowcaptionskip}{ -10 pt}
\centering
   \includegraphics[width=0.8\linewidth]{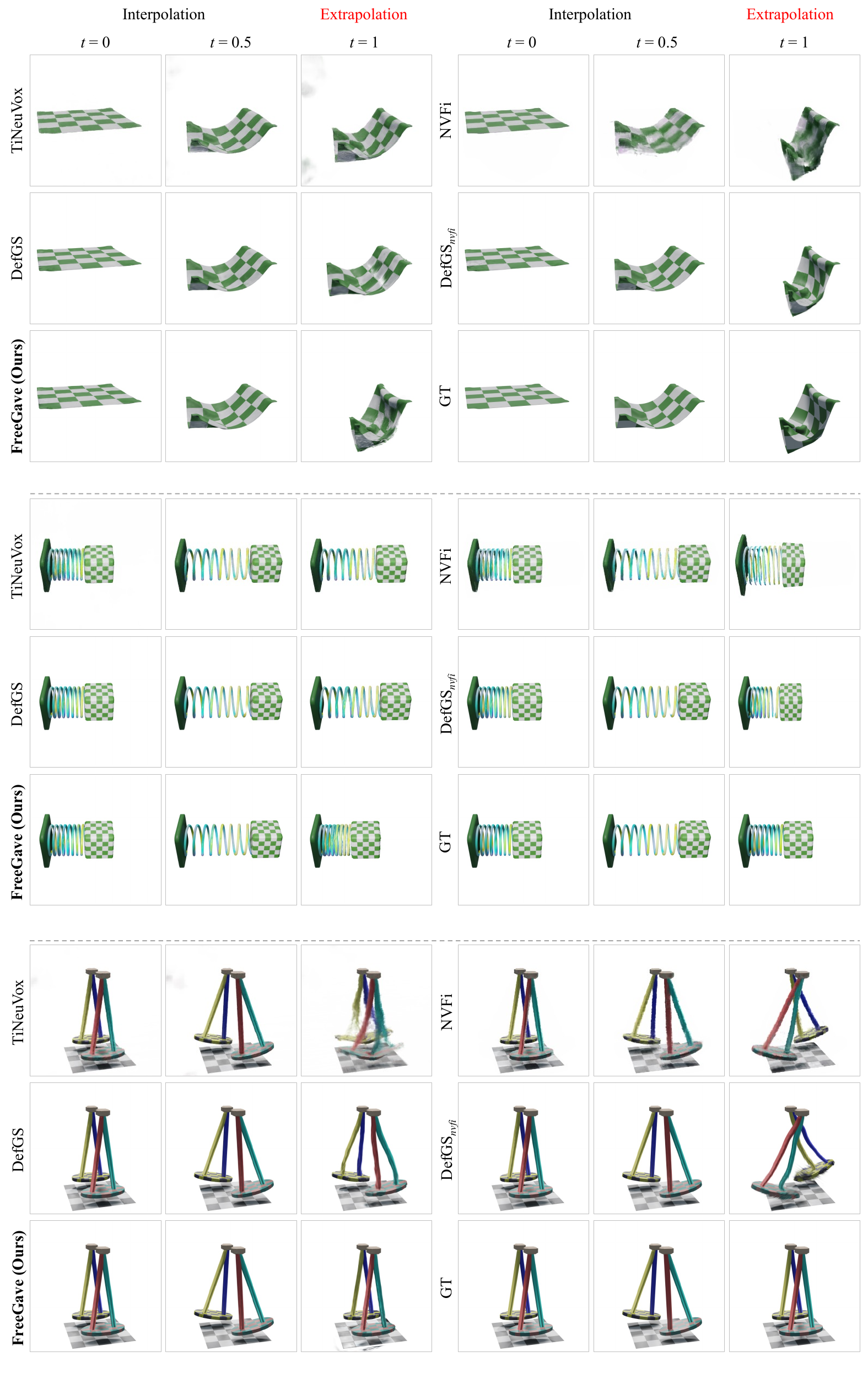}
\caption{Qualitative results for future frame extrapolation on ParticleNeRF Dataset.}
\label{fig:qual_res_app_particle2}
\end{figure*}

\begin{figure*}[t]
\setlength{\abovecaptionskip}{ 4. pt}
\setlength{\belowcaptionskip}{ -10 pt}
\centering
   \includegraphics[width=0.8\linewidth]{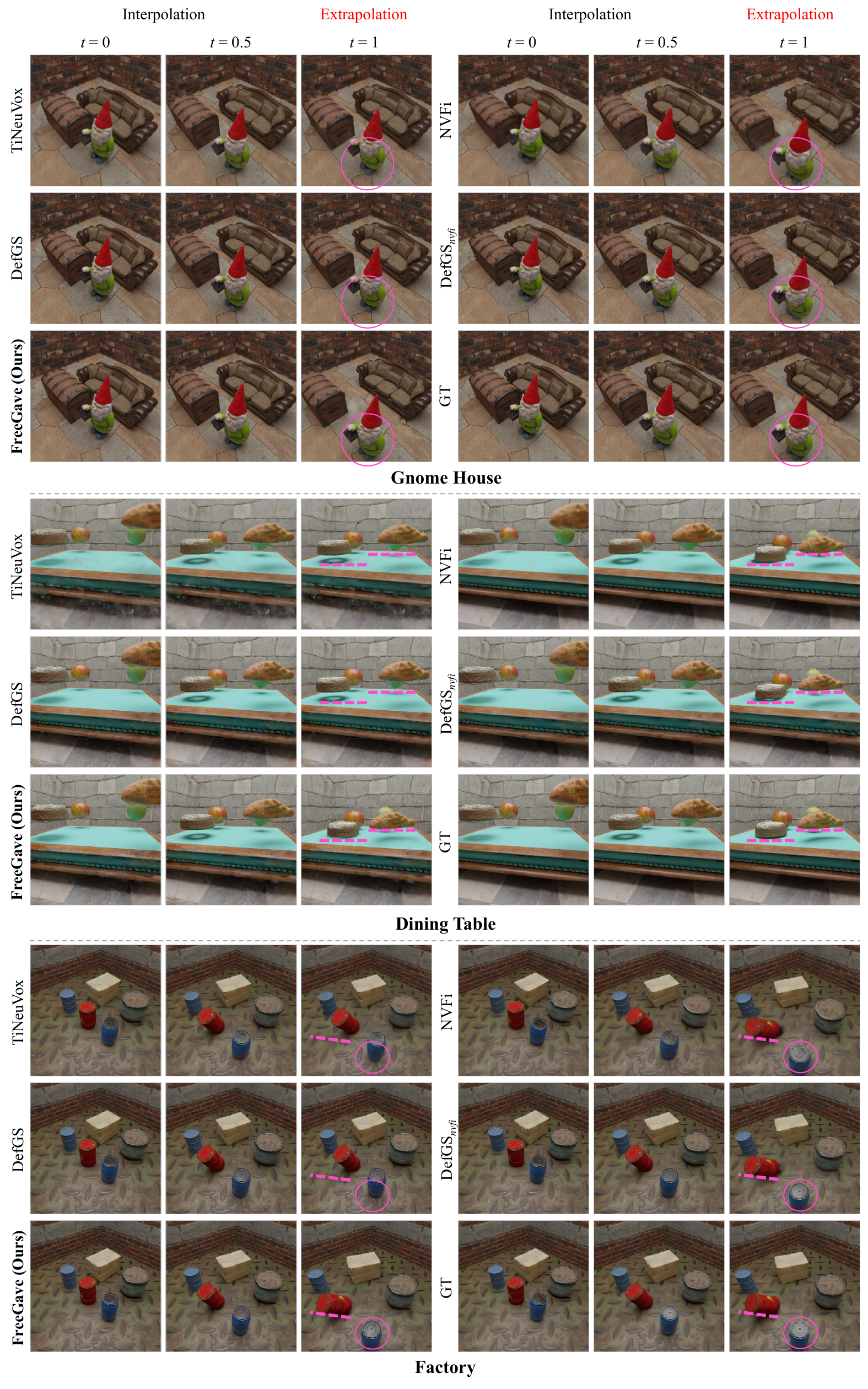}
\caption{Qualitative results for future frame extrapolation on Dynamic Indoor Scene Dataset.}
\label{fig:qual_res_app3}
\end{figure*}

\begin{figure*}[t]
\setlength{\abovecaptionskip}{ 4. pt}
\setlength{\belowcaptionskip}{ -10 pt}
\centering
   \includegraphics[width=0.8\linewidth]{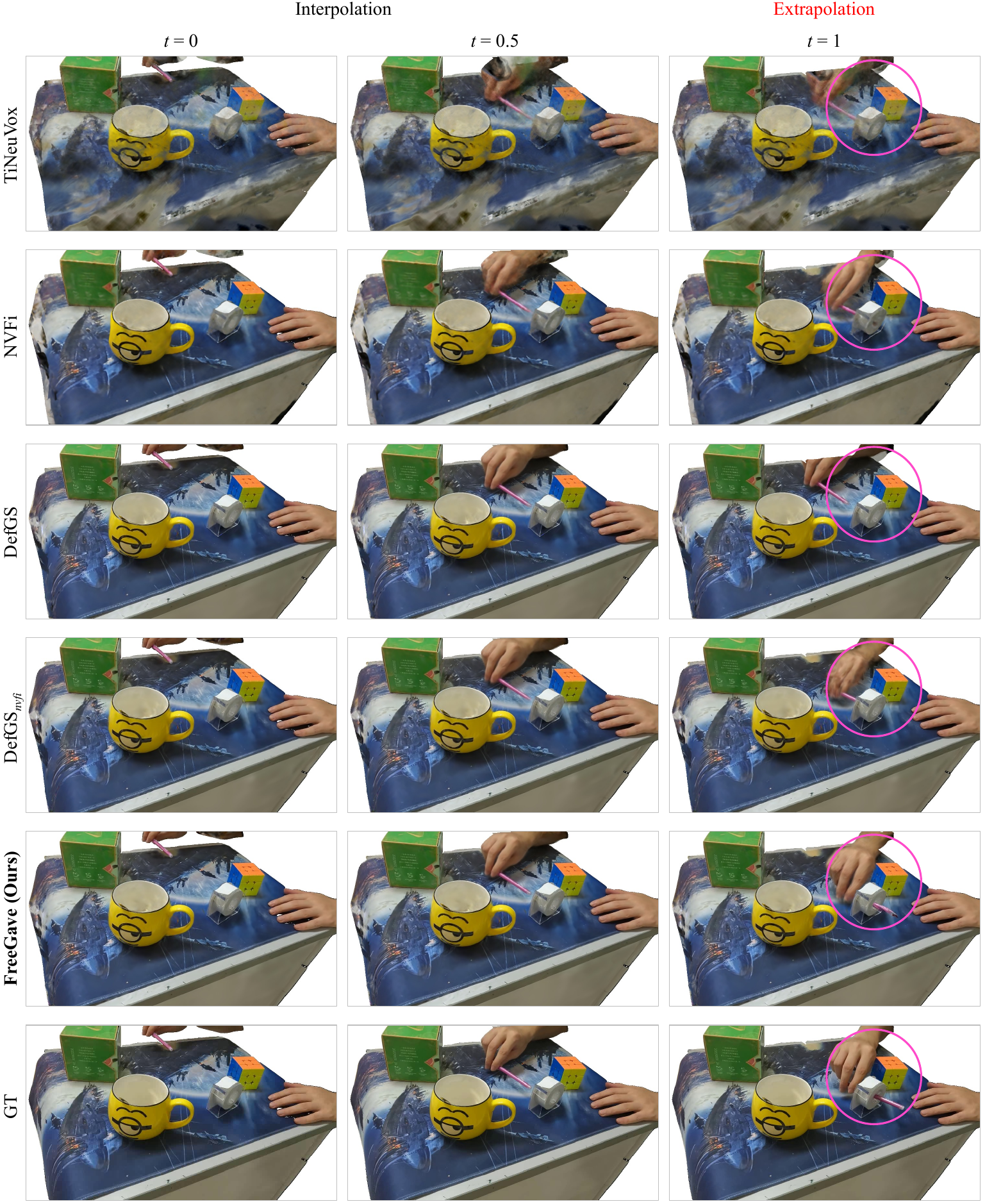}
\caption{Qualitative results for future frame extrapolation on ``Pen \& Tape 2" of \nickname{}-GoPro Dataset.}
\label{fig:qual_res_app4}
\end{figure*}

\begin{figure*}[t]
\setlength{\abovecaptionskip}{ 4. pt}
\setlength{\belowcaptionskip}{ -10 pt}
\centering
   \includegraphics[width=0.8\linewidth]{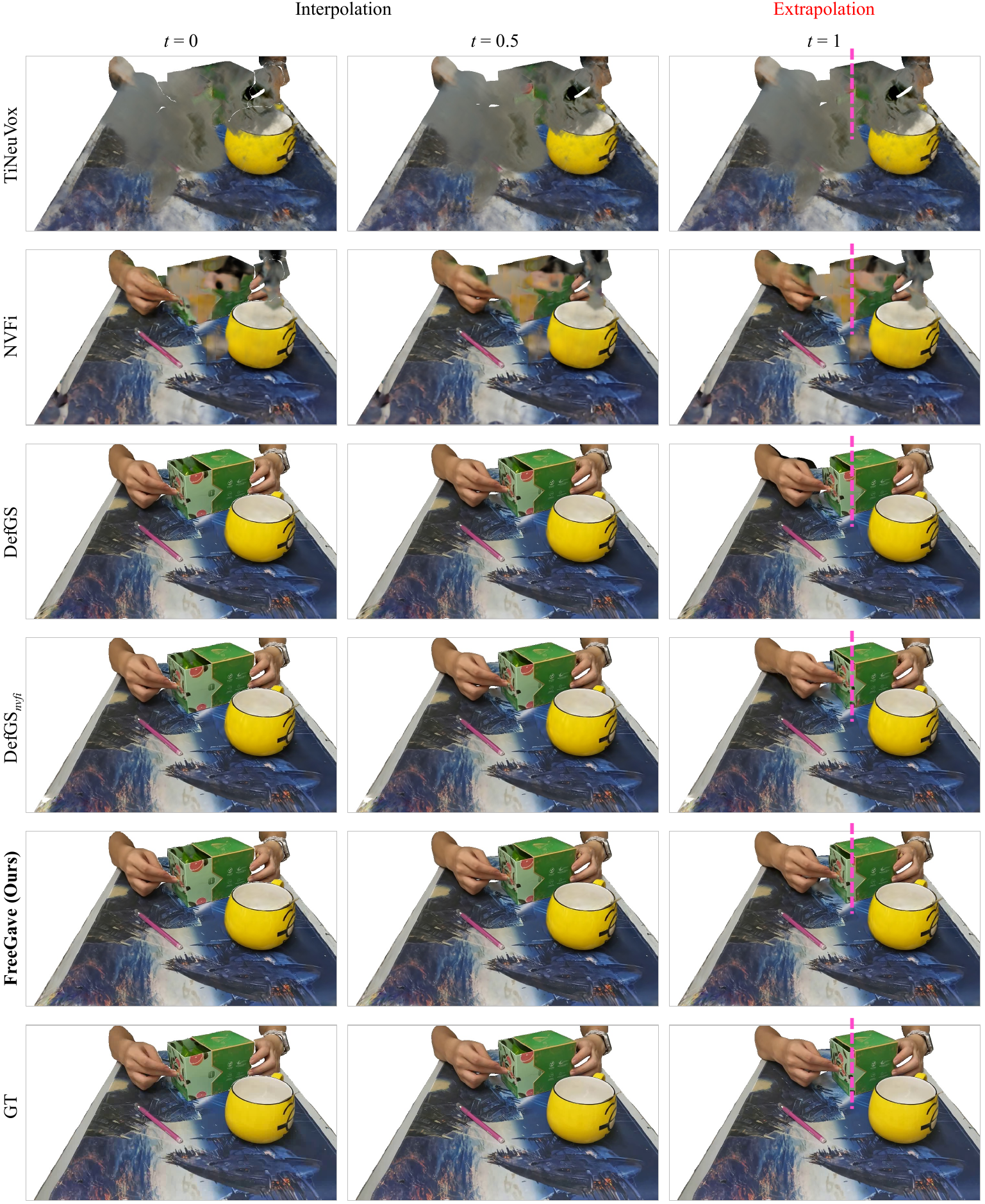}
\caption{Qualitative results for future frame extrapolation on ``Box" of \nickname{}-GoPro Dataset.}
\label{fig:qual_res_app5}
\end{figure*}

\begin{figure*}[t]
\setlength{\abovecaptionskip}{ 4. pt}
\setlength{\belowcaptionskip}{ -10 pt}
\centering
   \includegraphics[width=0.8\linewidth]{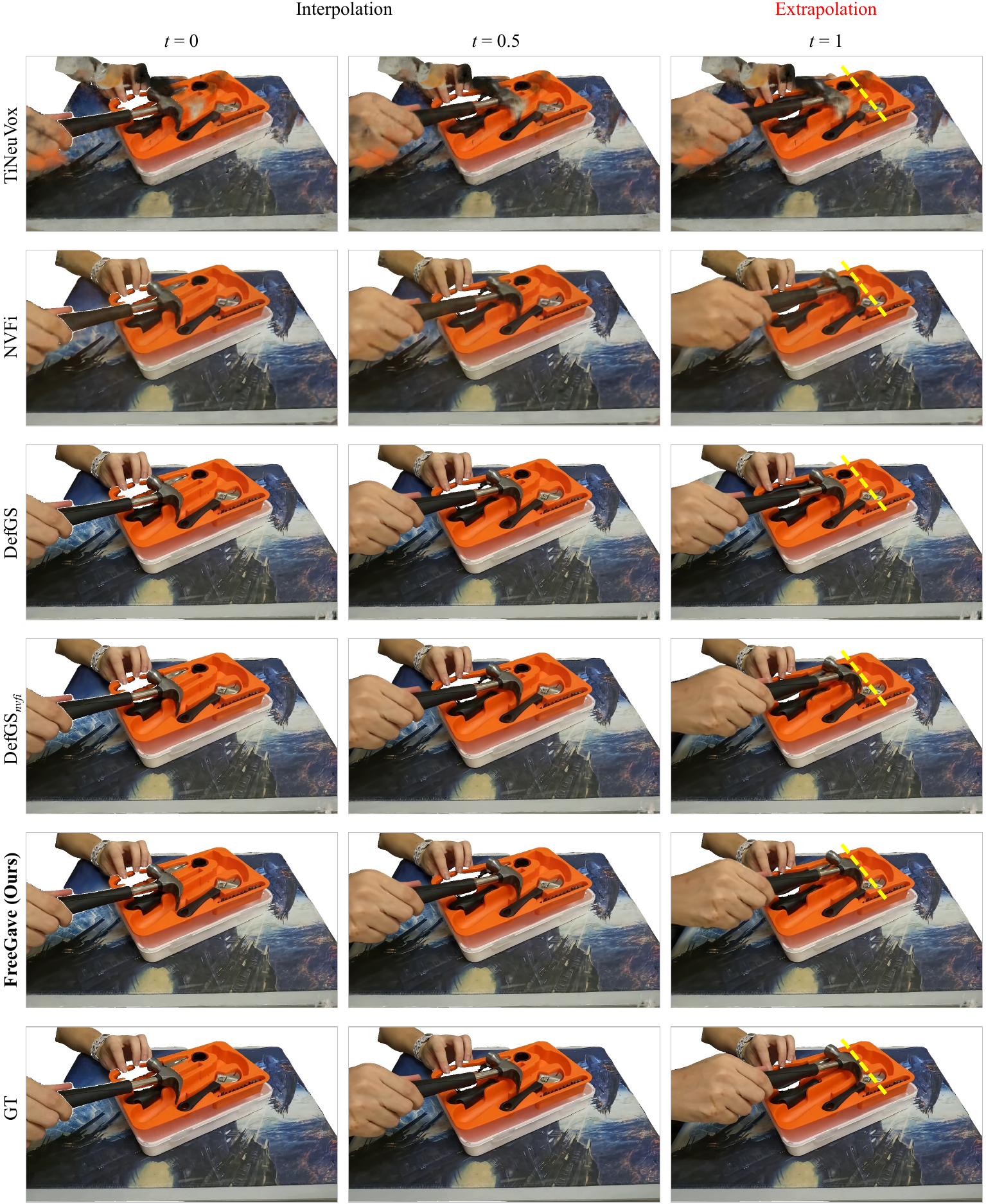}
\caption{Qualitative results for future frame extrapolation on ``Hammer" of \nickname{}-GoPro Dataset.}
\label{fig:qual_res_app6}
\end{figure*}

\begin{figure*}[t]
\setlength{\abovecaptionskip}{ 4. pt}
\setlength{\belowcaptionskip}{ -10 pt}
\centering
   \includegraphics[width=0.8\linewidth]{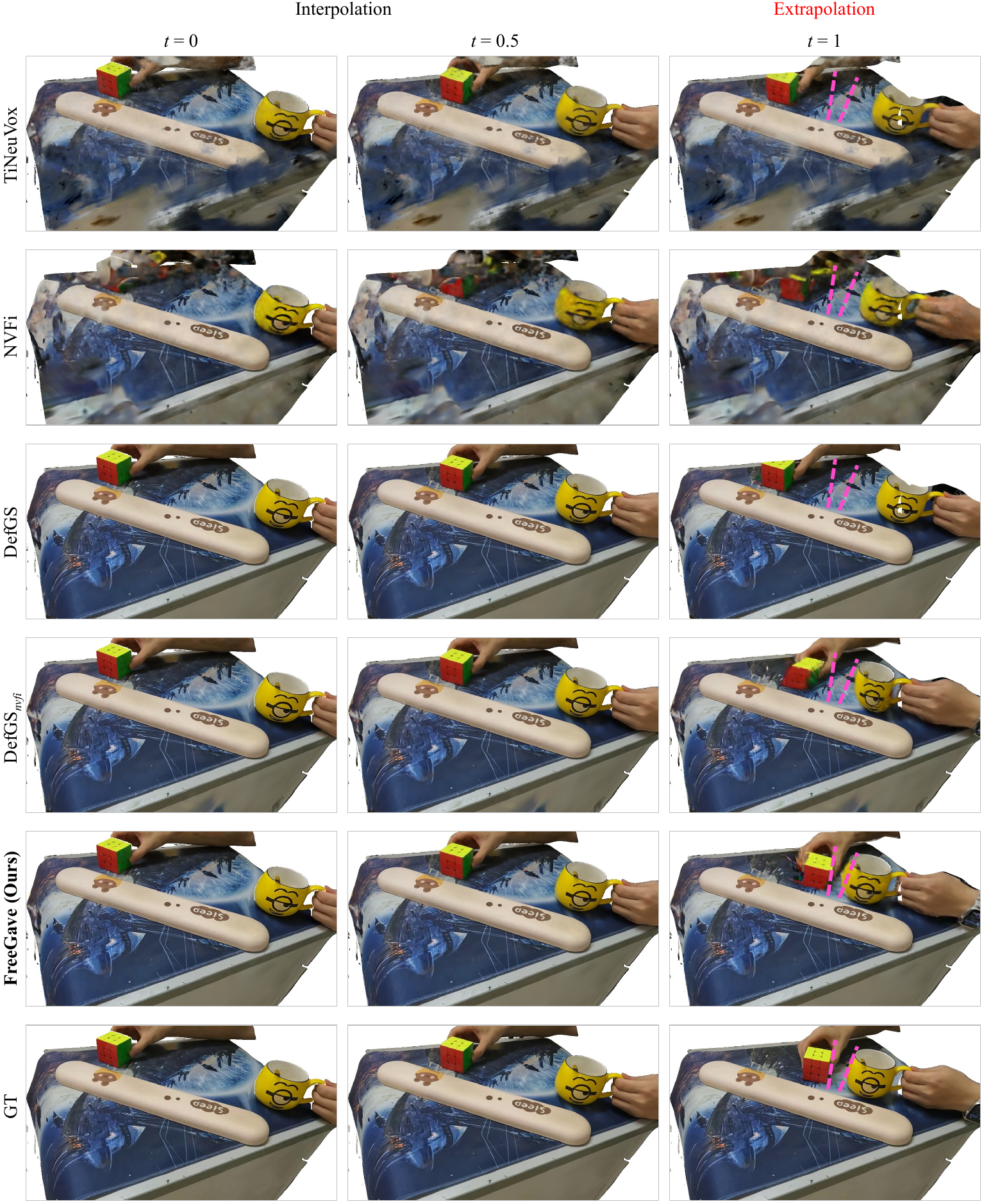}
\caption{Qualitative results for future frame extrapolation on ``Collision" of \nickname{}-GoPro Dataset.}
\label{fig:qual_res_app7}
\end{figure*}

\begin{figure*}[t]
\setlength{\abovecaptionskip}{ 4. pt}
\setlength{\belowcaptionskip}{ -10 pt}
\centering
   \includegraphics[width=0.8\linewidth]{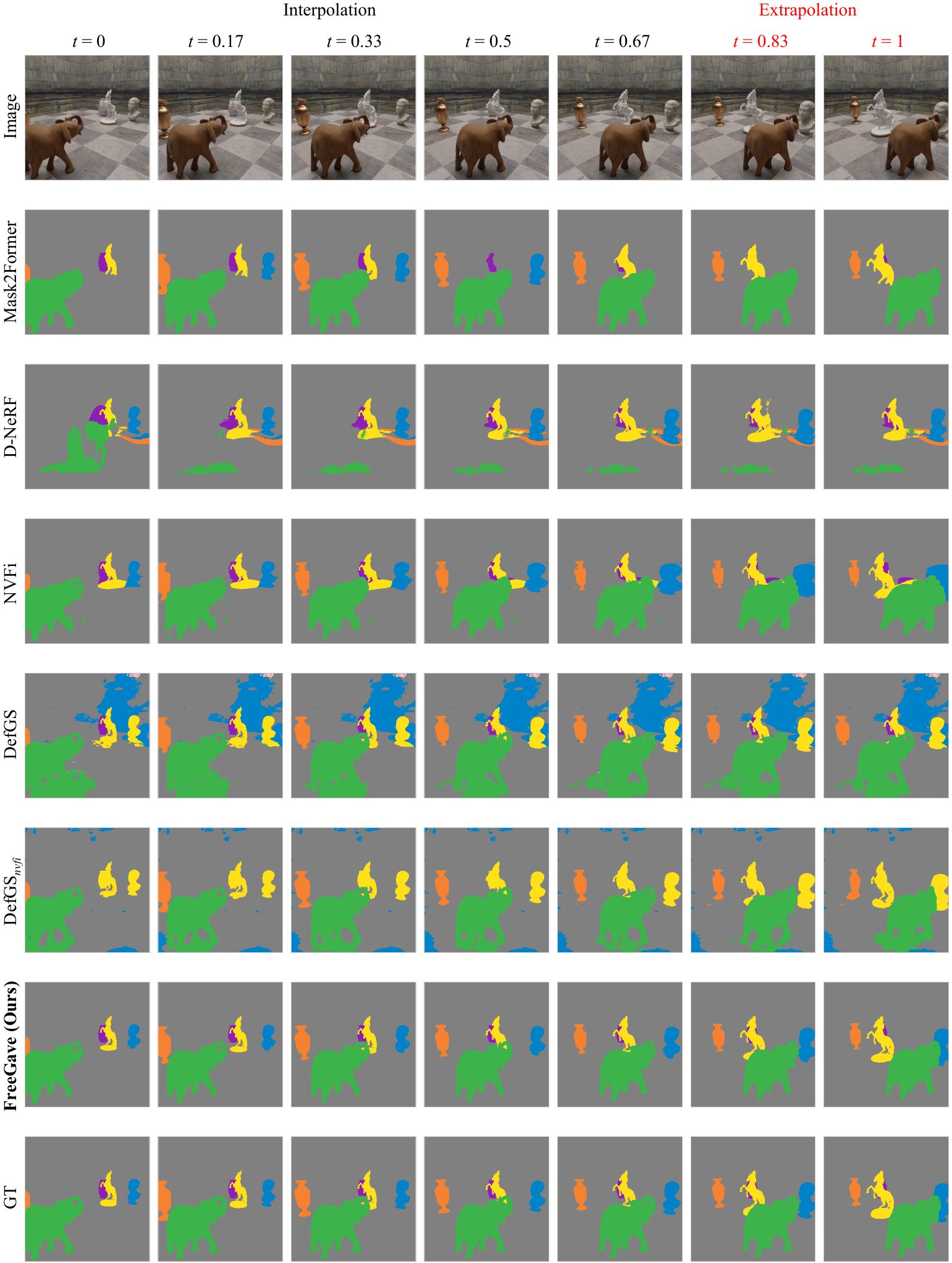}
\caption{Qualitative results for unsupervised motion segmentation on ``Chessboard" of Dynamic Indoor Scene Dataset.}
\label{fig:qual_res_app8}
\end{figure*}

\begin{figure*}[t]
\setlength{\abovecaptionskip}{ 4. pt}
\setlength{\belowcaptionskip}{ -10 pt}
\centering
   \includegraphics[width=0.8\linewidth]{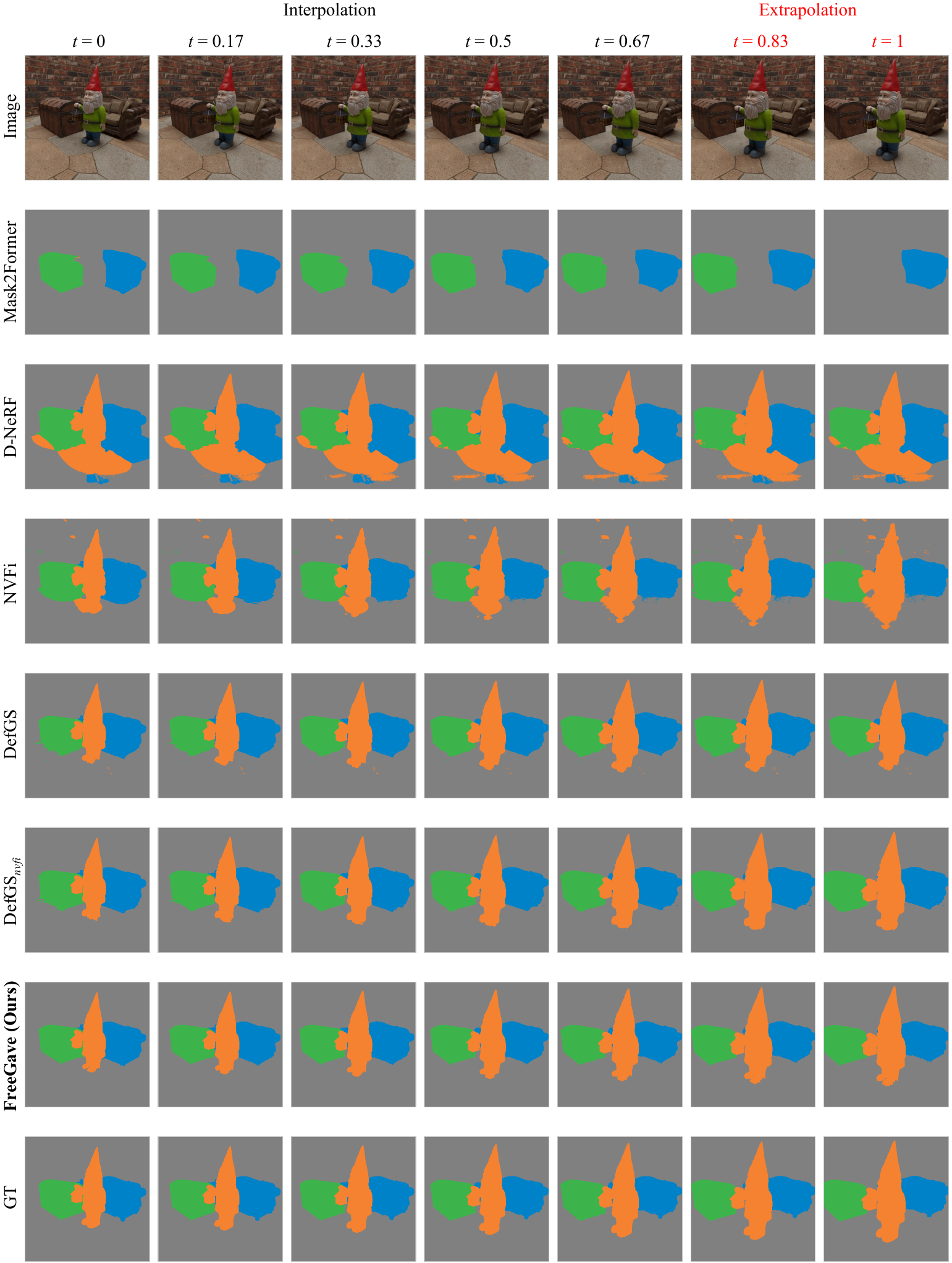}
\caption{Qualitative results for unsupervised motion segmentation on ``Gnome House" of Dynamic Indoor Scene Dataset.}
\label{fig:qual_res_app9}
\end{figure*}

\begin{figure*}[t]
\setlength{\abovecaptionskip}{ 4. pt}
\setlength{\belowcaptionskip}{ -10 pt}
\centering
   \includegraphics[width=0.8\linewidth]{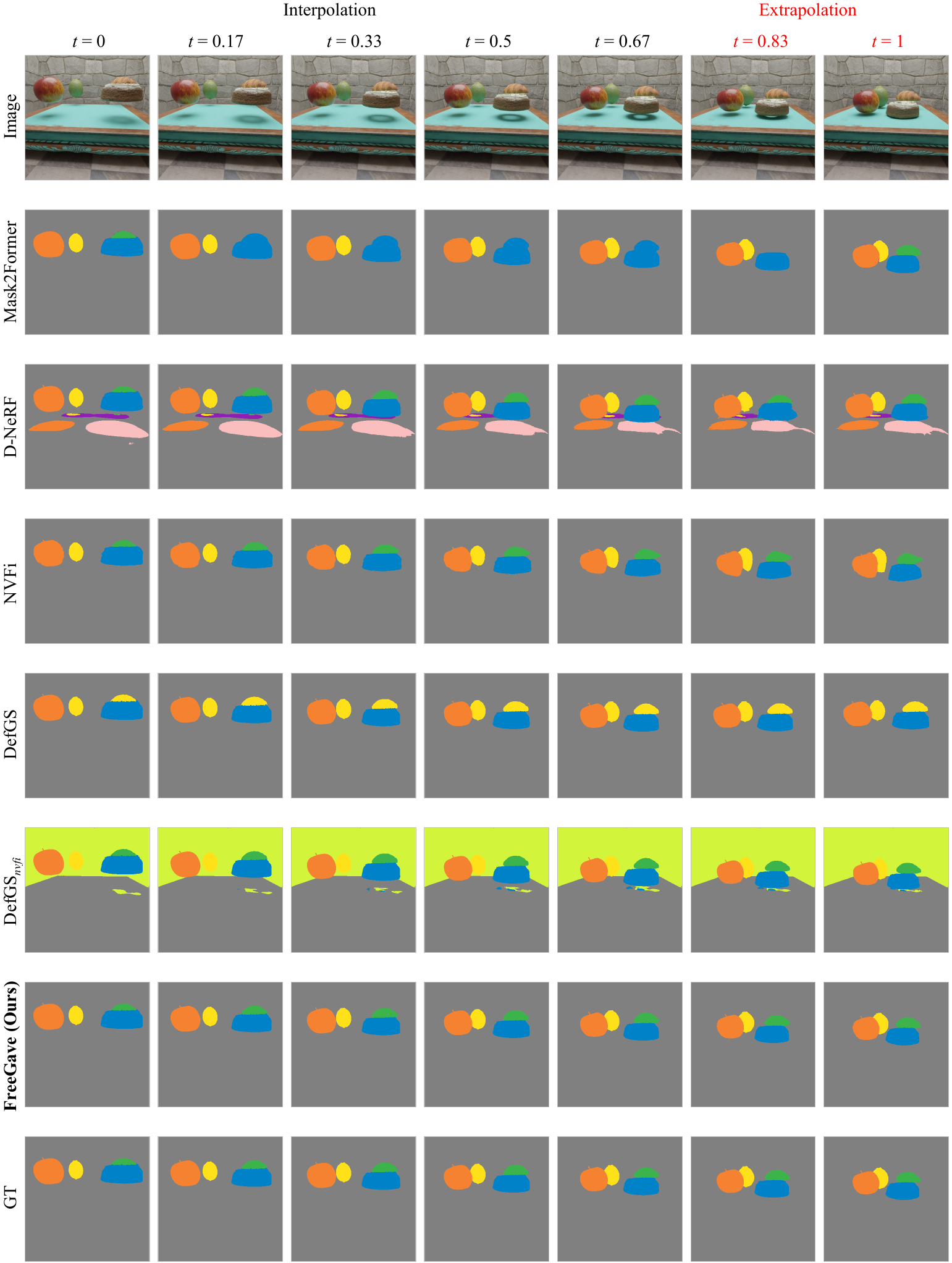}
\caption{Qualitative results for unsupervised motion segmentation on ``Dining Table" of Dynamic Indoor Scene Dataset.}
\label{fig:qual_res_app10}
\end{figure*}

\begin{figure*}[t]
\setlength{\abovecaptionskip}{ 4. pt}
\setlength{\belowcaptionskip}{ -10 pt}
\centering
   \includegraphics[width=0.8\linewidth]{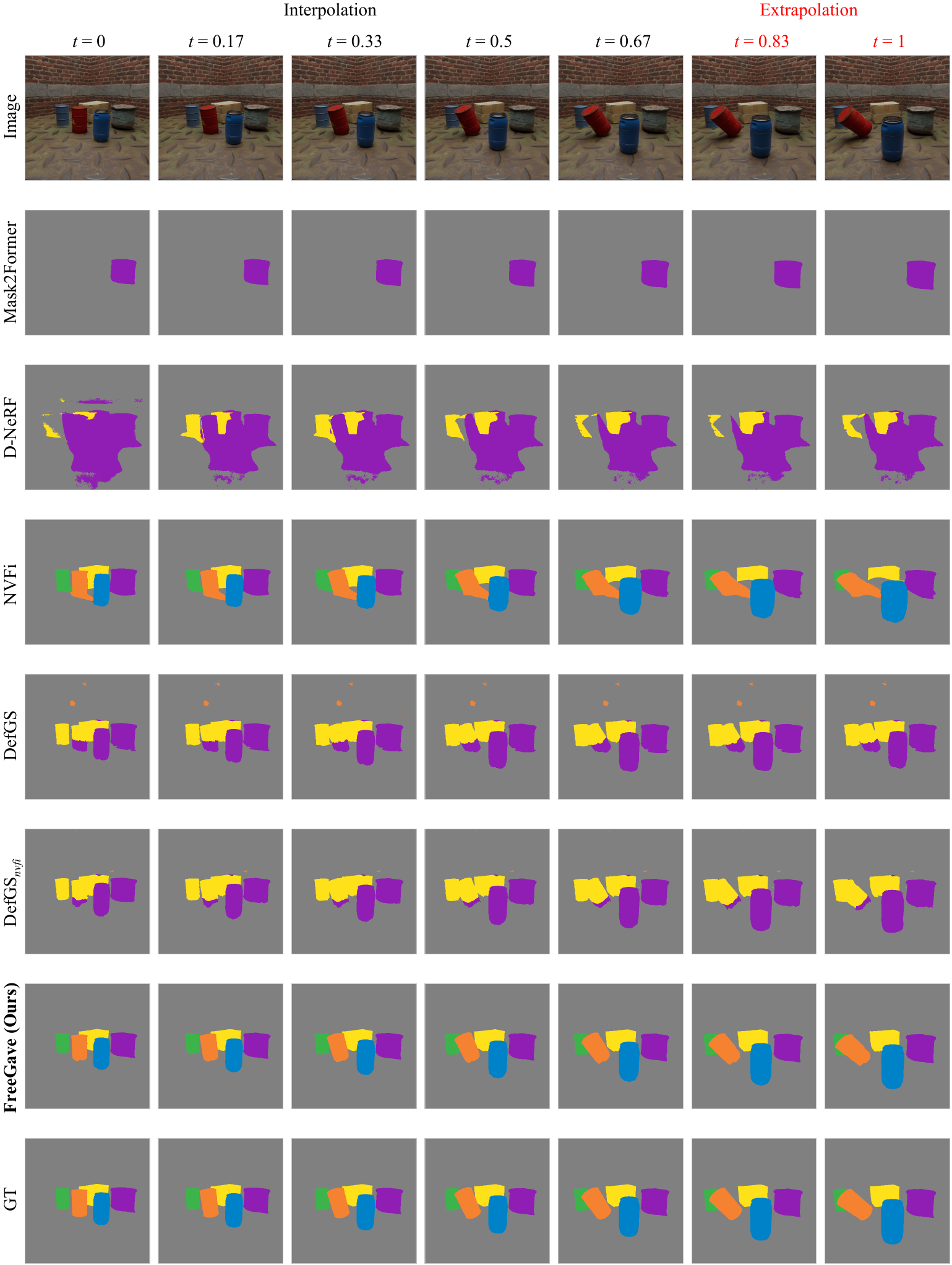}
\caption{Qualitative results for unsupervised motion segmentation on ``Factory" of Dynamic Indoor Scene Dataset.}
\label{fig:qual_res_app11}
\end{figure*}

We present additional qualitative results for future frame extrapolation in Figures \ref{fig:qual_res_app1}, \ref{fig:qual_res_app2}, \ref{fig:qual_res_app_particle1}, \ref{fig:qual_res_app_particle2}, \ref{fig:qual_res_app3}, \ref{fig:qual_res_app4}, \ref{fig:qual_res_app5}, \ref{fig:qual_res_app6},  and \ref{fig:qual_res_app7}. We also present additional qualitative results for scene decomposition in Figures \ref{fig:qual_res_app8}, \ref{fig:qual_res_app9}, \ref{fig:qual_res_app10}, and \ref{fig:qual_res_app11}.

\end{document}